%% file: main.tex
\documentclass{article}
\usepackage{kr}
\usepackage{enumitem}

\usepackage{times}
\usepackage{soul}
\usepackage{url}
\usepackage[hidelinks]{hyperref}
\usepackage[utf8]{inputenc}
\usepackage[small]{caption}
\usepackage{graphicx}
\usepackage{amsmath}
\usepackage{amsthm}
\usepackage{booktabs}
\usepackage{algorithm}
\usepackage{algorithmic}
\newtheorem{theorem}{Theorem}

\usepackage{amsmath}
\usepackage{amssymb}
\usepackage{mathtools}
\usepackage{amsthm}

\usepackage[capitalize,noabbrev]{cleveref}

\theoremstyle{plain}

\theoremstyle{definition}
\newtheorem{definition}[theorem]{Definition}

\theoremstyle{remark}

\usepackage[table,xcdraw]{xcolor}

\usepackage{dcolumn}
\newcolumntype{d}[1]{D{|}{\pm}{#1}}
\usepackage{subcaption}
\usepackage{booktabs}

\usepackage{listings}
\usepackage{caption}
\usepackage{etoolbox}

\usepackage{newfloat}
\DeclareFloatingEnvironment[name=Listing]{listing}

\DeclareCaptionFormat{algstyle}{%
  \hrule height 1.2pt \vspace{4pt}%
  \textbf{#1#2}#3\par\vspace{4pt}%
  \hrule height 0.6pt
}

\definecolor{pddlKeyword}{RGB}{0, 0, 130}
\definecolor{pddlLogic}{RGB}{90, 40, 90}
\definecolor{pddlVariable}{RGB}{120, 0, 0}

\lstdefinelanguage{PDDL}{
  morekeywords={
    :action,:parameters,:precondition,:effect,
    :domain,:requirements,:types,:predicates,:objects
  },
  morekeywords=[2]{and,not},
  sensitive=true,
  morecomment=[l]{;},
  morestring=[b]",
  alsoletter={:,-,?}, % treat :, -, ? as part of "words"
  keywordstyle=[1]\color{pddlKeyword}\bfseries,
  keywordstyle=[2]\color{pddlLogic}\bfseries, % ← different color for and/not
  stringstyle=\color{orange},
    literate=
    {?x}{{\color{pddlVariable}?x}}2
    {?y}{{\color{pddlVariable}?y}}2
    {?z}{{\color{pddlVariable}?z}}2
}

\newcommand{\tup}[1]{\langle #1 \rangle}
\usepackage{xspace} % xscape to insert a space after the macros when needed
\newcommand{\domain}{\texttt{simple}\xspace}
\newcommand{\bw}{\texttt{blocksworld}\xspace}
\newcommand{\bws}{\texttt{blocksworld-5b}\xspace}
\newcommand{\bwl}{\texttt{blocksworld-8b}\xspace}
\newcommand{\fr}{\texttt{ferry}\xspace}
\newcommand{\frs}{\texttt{ferry-5c}\xspace}
\newcommand{\frl}{\texttt{ferry-8c}\xspace}
\newcommand{\np}{\texttt{npuzzle}\xspace}
\newcommand{\nps}{\texttt{npuzzle-5t}\xspace}
\newcommand{\npl}{\texttt{npuzzle-8t}\xspace}
\newcommand{\mz}{\texttt{maze}\xspace}
\newcommand{\mzs}{\texttt{maze-5x5}\xspace}
\newcommand{\mzl}{\texttt{maze-7x7}\xspace}
\newcommand{\lgt}{\texttt{logistics}\xspace}
\newcommand{\lgts}{\texttt{logistics-3c}\xspace}
\newcommand{\lgtl}{\texttt{logistics-5c}\xspace}

\newcommand{\aone}{\texttt{a}}
\newcommand{\atwo}{\texttt{b}}
\newcommand{\athree}{\texttt{c}}

\newcommand{\pone}{\texttt{p}}
\newcommand{\ptwo}{\texttt{q}}
\newcommand{\pthree}{\texttt{r}}

\newcommand{\Omit}[1]{}

\usepackage{multirow}
\usepackage{stmaryrd}
\usepackage{placeins}  % add this in the preamble

\newcommand{\sift}{\textsc{sift}\xspace}
\newcommand{\strips}{\textsc{strips}\xspace}
\newcommand{\brasp}{\textsc{B-RASP}\xspace}

\newcommand{\del}[1]{\textcolor{red}{#1}}
\newcommand{\add}[1]{\textcolor{lightgray}{#1}}

\newcommand{\trace}{\tau=(a_0,\ldots,a_{n-1})}
\newcommand{\setupinit}{\text{\emph{init-setup}~}}
\newcommand{\setupend}{\text{\emph{end-setup}~}}

\newcommand{\carlos}[1]{{\textcolor{blue}{Carlos: #1}}}
\newcommand{\VG}[1]{{\textcolor{brown}{Vicen\c{c}: #1}}}
\newcommand{\hector}[1]{{\textcolor{red}{Hector: #1}}}

\title{From Next Token  Prediction to (STRIPS) World Models}

\author{%
Carlos Núñez-Molina$^1$,
Vicenç Gómez$^2$,
Hector Geffner$^1$ \\
\affiliations
$^1$RWTH Aachen University, Germany\\
$^2$Universitat Pompeu Fabra, Spain\\
\emails
\{carlos.nunez,hector.geffner\}@ml.rwth-aachen.de,
vicen.gomez@upf.edu
}

\begin{document}
\maketitle
\begin{abstract}
We study whether next-token prediction can yield world models that truly support planning, in a controlled symbolic setting where propositional \strips action models are learned from action traces alone and correctness can be evaluated exactly.
We introduce two architectures. The first is the \strips Transformer, a symbolically aligned model grounded in theoretical results linking transformers and the formal language structure of \strips domains. The second is a standard transformer architecture without explicit symbolic structure built in, for which we study different positional encoding schemes and attention aggregation mechanisms.
We evaluate both architectures on five classical planning domains, measuring training accuracy, generalization, and planning performance across domains and problem sizes. Interestingly, both approaches can be used to produce models that support planning with off-the-shelf \strips planners over exponentially many unseen initial states and goals. Although the \strips Transformer incorporates a strong symbolic inductive bias, it is harder to optimize and requires larger datasets to generalize reliably. In contrast, a standard transformer with stick-breaking attention achieves near-perfect training accuracy and strong generalization. Finally, standard transformers without stick-breaking attention do not generalize to long traces, whereas a  symbolic \strips model extracted from a transformer trained on shorter traces does.
% but  as well as the STRIPS Transformer, either the standard attention mechanism and positional encodings have to be replaced by stick-breaking attention, or a symbolic STRIPS model needs to be extracted from the learned Transformer and used instead.
\end{abstract}

% TODO -> resumir abstract a 200 palabras y un único párrafo!
% Guidelines:
%The abstract should be a concise, one-paragraph summary describing the general thesis and conclusion of your paper. A reader should be able to learn the purpose of the paper and the reason for its importance from the abstract. The abstract should be no more than 200 words long.

\input{introduction.tex}

\section{Related Work}
\input{related.tex}

\section{Background}
\input{background.tex}

\section{Learning Task}
\input{task.tex}

\section{\strips Transformer}

\input{architecture.tex}

\section{Stick-Breaking Transformer}
\input{sb_transformer.tex}
\input{learning.tex}

\section{Planning}

\input{planning.tex}

\section{Experiments}
\input{experiments.tex}

\section{Results}
\input{results.tex}

\section{Summary}
\input{conclusion.tex}

\section*{Acknowledgements}
This research has been partially supported by the Alexander von Humboldt Foundation with funds from the Federal Ministry for Education and Research, by the European Research Council (ERC), Grant agreement No.
885107, and by the Excellence Strategy of the Federal Government and the NRW state, Germany. This publication
is also part of the action CNS2022-136178 financed by
MCIN/AEI/10.13039/501100011033 and by the European Union Next Generation EU/PRTR.

\section*{AI Declaration}
During the preparation of this work the authors used OpenAI's ChatGPT  in order to enhance readability and perform grammar checking. In addition, ChatGPT Codex was used to support code development. All outputs generated by these tools were carefully reviewed by the authors, and modified where necessary. The authors take full responsibility for the content of the publication.

%% The file kr.bst is a bibliography style file for BibTeX 0.99c

\bibliographystyle{kr}
\bibliography{references}

\newpage
\onecolumn

\appendix
\part*{Appendix}
\input{brasp.tex}

\input{appendix}

\end{document}

%% file: introduction.tex
\section{Introduction}

A number of recent works have considered whether LLMs, and in particular transformer
architectures, can learn world models as a result of learning to predict the
next token autoregressively \cite{world-models1,world-models2}.
This question is central because genuine understanding requires the ability to make
meaningful predictions from an internal model of the world, and it remains unclear
whether LLMs actually acquire such models, rather than relying solely on surface
statistical regularities \cite{mitchell:world-models}.

%A number of recent works have  considered whether LLMs, and in particular, transformer
%architectures,   can learn world models as a result of  learning  to predict the
%next token ``autoregressively'' \cite{world-models1,world-models2}.
%The question  is important because true understanding involves the ability to make
%meaningful predictions from a model, and it is far from  clear whether LLMs
%exhibit true understanding or learn world models \cite{mitchell:world-models}. 

% vicenc: added chess ref
%Kenneth Li and co-authors trained a transformer network,
%called “OthelloGPT,” to play Othello~\cite{li:world-models}. For verifying 
%if a world model was learned, the authors determined whether the embedding of the
%board cells reflected indeed the state of the cell
In order to test the hypothesis that transformer networks can learn world models when 
predicting the next token, several works have examined controlled game environments such
as Chess and Othello \cite{Toshniwal_Wiseman_Livescu_Gimpel_2022,li:world-models,nanda_othello_2023}.
\Omit{
In the
Othello case, Kenneth Li and co-authors trained a transformer network, called “OthelloGPT”,
using next-token prediction alone and evaluated whether the embeddings associated with board
cells reflected the underlying game state, namely whether a cell was empty or contained a
white or black piece. Interestingly, the cell embeddings did not reveal the color of the
pieces directly, but instead whether the piece belonged to the player who moved last or to
the other player \cite{nanda_othello_2023}.
}
Learning world models that support planning, however, requires learning   both a representation of states
and a model of how those states evolve, and while transformers have been shown to learn latent state representations, these representations have
not been found to be good enough to support  planning \cite{world-models2}.

%The problem of  learning world models from next token prediction
%requires learning a state representation and the ways in which
%the  states change.  Transformers have been shown to be capable of learning such state representations,
%yet these  representations have not been found to be good enough for planning~\cite{world-models2}.

% \footnote{ %%% Hector:
%% This relation is not evident; in \cite{world-models3}, world models are not learned accurately,
%% 5 in value-equivalence, it's not a matter of learning the model accurately; it's
%% that the world model is not needed for the task, and could be changed while preserving V*.
% 
% This relates closely to the notion of value-equivalent action models \cite{grimm2020value}, i.e.,
% approximate action models learned to solve a particular task, and which do not represent the 
% dynamics of the underlying system in a faithful manner.
% }
%Transformers have been shown to be capable of learning such state representations, yet these  representations are often the result of shortcuts and heuristics that do not capture the true world model, and reflect instead shortcuts and heuristics \cite{world-models3}.
% We don't cite the work "world-models4" as it focuses on SSMs instead of transformers

This work  is aimed at relating the tasks of next token prediction
and world-model learning using transformers in the very concrete
setting of propositional \strips world models.
Propositional \strips world or action models have
been standard in AI for many years
\cite{russell:book,ghallab:book,geffner:book}, 
and they represent deterministic MDPs in compact form,
where the states are given by  sets  of atoms,
and  actions add and delete  atoms, provided that their  preconditions
are true. An atom in \strips stands for a Boolean variable,
and an atom in the state, which is a collection of atoms,
represents a Boolean variable that is true in the state.

For learning \strips models from action traces, we introduce two specialized architectures: the \strips Transformer and the Stick-Breaking (SB) Transformer. Both are motivated by recent results connecting hard-attention transformers and the B-RASP formalism to star-free languages \cite{b-rasp}.\footnote{For the connections with B-RASP, see the Appendix.} The \strips Transformer incorporates an explicit symbolic alignment to \strips structure, whereas the SB Transformer retains a standard decoder-style architecture but removes positional encodings and replaces softmax with stick-breaking attention, a  sequential normalization mechanism~\cite{tan2025scaling}. Both models learn \strips action models from sequences of applicable (positive) and non-applicable (negative) actions obtained from random walks together with observations of action applicability.
We show experimentally that both architectures generalize compositionally over long horizons and support planning over exponentially many unseen initial states and goals using off-the-shelf \strips planners.

The paper is organized as follows. We first review related work and background on \strips and transformers. Then, we introduce the learning task, describe the \strips and SB transformers, their training process, and how \strips models are extracted and used for planning. Finally, we present the results of our experiments and summarize our findings.\footnote{
Code and data at \url{github.com/TheAeryan/strips-transformer}.
%Extended version with Appendix at  \url{arxiv.org/abs/2509.13389}.
}

% End Omit

%% file: related.tex
% \section{Related Work}

\Omit{
  This work relates to three research threads: understanding transformers
  in terms of the formal languages they can recognize, learning \strips planning
  models, and learning world-models from next token prediction. We discussed the last thread
  above, and focus below on the first two.}

% This work relates to several research threads.

\medskip

\noindent \textbf{Transformers and World Models}. 
% World models are used to  predict future states and rewards  from observations and actions, primarily to support policy learning, imagination, or planning in model-based RL settings \cite{david_schmidhuber_2018,zhu2023value}.
% In this context,
Transformer architectures trained for next-token prediction in the RL setting have motivated the question of  whether  sequence models can
internalize aspects of environment dynamics. The common methodology in this line of work is to \emph{probe} trained transformers in controlled environments to test whether latent state or causal structure can be inferred from the learned token embeddings \cite{li:world-models,Toshniwal_Wiseman_Livescu_Gimpel_2022,nanda_othello_2023,rohekar2025a,vafa2025what}. By contrast, other  approaches propose modified architectures or structured inductive biases to improve the usability of learned world models for planning and long-horizon reasoning, often by introducing explicit structure, abstraction, or symbolic components \cite{pmlr-v139-zhang21x,feng2025graph,cano2025neurosymbolic}.
%% Missing a sentence here that relates our work
%However, these approaches do not recover exact symbolic transition models from action-only traces that can be used for planning.
None of these methods, however, are  aimed at  recovering  exact symbolic models that can be used with off-the-shelf classical planners,
as the methods proposed in this work. 

\Omit{ % Vicenc: previous paragraph on transformers and world models with less citations
    \noindent \textbf{Transformers and World Models}. As reviewed in the introduction, 
    transformer architectures used to learn suitable policies have been shown
    to learn approximate world models~\cite{world-models1,Toshniwal_Wiseman_Livescu_Gimpel_2022,li:world-models}.
    Indeed, the ability to achieve goals in one step implies the ability to predict one-step changes~\cite{}.
    Yet the representations learned for achieving given goals do not appear
    to yield world models that are good enough for planning,
    while generating goal-directed traces too different from those seen at training~\cite{vafa2025what}.
}

\medskip

\noindent \textbf{Transformers and RASP languages}.
A simple programming language, called RASP, was introduced to capture the computations
performed by transformers  at a more abstract level \cite{rasp},
and a Boolean  version of RASP, called B-RASP, has been  proposed  to characterize these
computations in terms of a class of  formal languages, specifically, the star-free languages \cite{b-rasp}.
Interestingly, it has been shown that the language of valid action traces induced by a \strips planning domain
is itself star-free \cite{Lin_Bercher_2022}, which suggests that B-RASP programs, and therefore hard-attention transformers,
are expressive enough to recognize such domains. We build on this result in this work.
%General B-RASP programs can  be compiled into hard-attention
%transformers, and we build on this  result in this work.

\medskip

\noindent \textbf{\strips Model Learning.} Symbolic algorithms for learning
\strips models from traces composed of actions and states
have been considered in a number of works \cite{zhuo2013action,aineto2019learning,lamanna2025lifted}.
Other symbolic approaches, such as LOCM and SIFT, learn models from action traces alone
\cite{locm,sift}, while others rely on action names together with the structure of
the state graph \cite{bonet:ecai2020,ivan:kr2021}. 
Fewer works address \strips model learning in the deep learning setting
\cite{asai:jair,xi2024neuro}, and none of them learn from action traces alone,
as done in this work.

\medskip
\noindent \textbf{Pre-trained LLMs and Planning.} Large language models (LLMs) have been increasingly explored as tools for planning. Early work investigated whether LLMs
can  act directly as planners by generating complete plans or plan fragments from task descriptions \cite{silver2022pddl,kambhampati2024position,pallagani2023plansformer}. A complementary line of work incorporates LLMs into the planning process itself, using them to guide or structure combinatorial search \cite{hao2023reasoning,yao2023tree,besta2024graph,katz2024thought}. 
% Carlos: removed paragraph below for summarization
%Among these, Thought of Search (ToS) \cite{katz2024thought} queries a pre-trained LLM to synthesize, for a given domain, an explicit successor function and goal test over an explicit state representation, and then plugs these components into standard search procedures such as BFS. 

\medskip

Our work is most closely related to approaches that leverage LLMs to produce planning-related components, such as domain models or generalized planning programs, which are then input to classical planners or executed directly \cite{NEURIPS2023_f9f54762,oswald2024large,gestrin2024towards,silver}. Unlike these works, which rely on pre-trained LLMs, we learn a symbolic transition model from action-only traces, and evaluate generalization over exponentially many unseen initial states and goals.

%% file: background.tex
\subsection{\strips Models}

\Omit{
  Propositional \strips planning problems are factored, deterministic MDPs where the state is given by the value of
a set of Boolean variables, and the actions change the value of these variables provided its preconditions hold
\cite{russell:book,ghallab:book,geffner:book}. 
Actions are regarded as not applicable in a state when its preconditions do not hold.
% , or alternatively, actions can be regarded as not applicable when then they do not affect the state. 
}

Propositional (grounded) \strips problems  $P$ are  expressed as tuples of the form $P = \tup{F,A,I,G}$
where $F$ is a set of atoms or propositions, $A$ is set of actions, 
$I \subseteq F$ is a subset of atoms   defining the initial situation, 
and  $G \subseteq  F$ is a subset of atoms defining the goal \cite{russell:book,ghallab:book,geffner:book}. 
The actions $a \in A$ are  defined in terms of  three subsets of atoms  from $F$:
the precondition list $\mathrm{pre}(a)$, the add list $\mathrm{add}(a)$, and the delete list $\mathrm{del}(a)$.

A problem $P=\tup{F,A,I,G}$ defines a state model (deterministic MDP)  $S(P)=\tup{S,s_0,S_G,A,A(\cdot),f}$ where
the states $s \in S$ are subsets of $F$, the initial state $s_0$ is $I$,
the goal states $s \in S_G$ include all the goal atoms,
$A(s)$ is the set of applicable actions in $s$ given by those in $A$ whose preconditions $\mathrm{pre}(a)$ are all
true in $s$, and a deterministic state-transition function $f(a,s)$ that maps $s$ into the state
$s' = (s \cup \mathrm{add}(a)) \setminus \mathrm{del}(a)$ when $a \in A(s)$.

The atoms $p$ in a state $s$ represent the subset of atoms in~$F$ that are true in $s$.
An action sequence $\trace$ over~$P$ is a sequence of actions from $A$.
The action sequence  is applicable in a state $s$ in $S(P)$ if there is a sequence
of states $s_0, \ldots, s_{n}$ in $S(P)$ such that each action $a_i$ is applicable in $s_i$,
and $s_{i+1}$ is the state that results from action $a_i$ in $s_i$, i.e., $s_{i+1}= f(a_i,s_i)$,
$i=0, \ldots, n-1$.  The action sequence $\tau$ is a plan for $P$ if it is applicable in
the initial state of $P$ and the resulting state $s_{n}$ is a goal state.

A  propositional \strips \emph{domain} or \emph{model} $M$ is given
by the set of atoms $F$  and the set of actions $A$, as $M=\tup{F,A}$.
A model is compatible with a number of \emph{problem instances}
$P=\tup{F,A,I,G}$, all sharing the same atoms and the same  actions but
differing in the initial situation $I$ or goal $G$. The number of instances
$P$ of a given model $M$ is exponential in the number of atoms $F$ in $M$.

% \section{Attention Mechanisms}

\subsection{Transformers}

\label{sec:attention-background}

% Carlos: added info about positional encodings and attention heads
We will use transformers for learning propositional \strips models from action sequences.
%Transformers~\cite{transformers} are suited for processing sequences of tokens $\tau = \tup{b_1,\ldots, b_n}$, outputting an updated sequence $\tau' = \tup{b'_1,\ldots, b'_n}$.
Transformers~\cite{transformers} are suited for processing sequences of tokens $\tau=\langle b_0,\ldots,b_{n-1}\rangle$, outputting an updated
sequence $\tau'=\langle b'_0,\ldots,b'_{n-1}\rangle$.
Input tokens $b_i$ are represented by real vectors often containing information about their position~$i$,
in which case they are called \textit{positional encodings/embeddings}.
Token embeddings (positional or not) are updated across one or more layers.
Transformer layers contain feed-forward nets and perform the operation known as \emph{self-attention},
where tokens receive signals from other tokens.
Often, each layer will perform several self-attention computations in parallel (each as a separate \textit{attention head}),
in order to extract different types of relations between tokens.

% Hector's version
\iffalse
We will use transformers for learning propositional \strips models from action sequences.
Transformers~\cite{transformers} are indeed suited for processing sequences of tokens $\tau = \tup{b_1,\ldots, b_n}$,
where each token is given by a real vector of a fixed dimension $d$, producing an updated
sequence of tokens $\tau' = \tup{b'_1,\ldots, b'_n}$.
%of vectors with the same size. % Carlos: not necessarily of the same size
In a transformer, each token is updated by a feed-forward net, and tokens
receive signals from other tokens via the mechanism known as \emph{self-attention}.
\fi

% Transformers process an input sequence by means of \emph{self-attention}.  
Given an input sequence  of length $n$, self-attention computes, for each position $i$, a weighted combination of
representations at positions $j$ in the same sequence. %
In its simplest  \emph{scalar} form, each position $i$ is associated with a real \emph{query} $Q(i)$, \emph{key} $K(i)$, and \emph{value} $V(i)$.  
The compatibility between positions $i$ and $j$ is measured by a score %\hector{need to talk about Heads here}
\[
S(i,j) = Q(i) \cdot K(j).
\]
To ensure that predictions at position $i$ depend only on preceding positions, a \emph{strict future mask} is applied by setting $S(i,j)=0$ whenever $j \ge i$.

The output at position $i$ is obtained by aggregating the values $V(j)$ according to weights derived from the masked scores $S(i,j)$. Different attention mechanisms correspond to different ways of selecting or normalizing these weights.

% \paragraph{Masked hard attention.}
In \emph{masked hard attention}, each position $i$ selects a single preceding position $j<i$ with maximal score, breaking ties by choosing the largest such index.
This form of attention has been studied formally in the context of B-RASP,  where it is shown that the Boolean operations of transformers
with masked hard attention can be captured in symbolic B-RASP programs \cite{b-rasp}.

\Omit{
Formally, letting
\[
j^*(i) = \max \left\{\, j<i \mid S(i,j) \text{ is maximal among } j<i \,\right\},
\]
the output at position $i$ is $V(j^*(i))$, and a default value is used if no such $j$ exists.  
Thus, each position attends to exactly one previous position.  
This mechanism corresponds to the attention operators used in B-RASP programs~\cite{b-rasp} and naturally captures recency-based dependencies.
}

% \paragraph{Stick-breaking attention.}
\emph{Stick-breaking attention} \cite{tan2025scaling} provides a differentiable mechanism that approximates masked hard attention while preserving its recency bias.  
Given masked scores $S(i,j)\in[0,1]$, it defines attention weights as
\[
S'(i,j) = S(i,j)\,\prod_{k=j+1}^{i-1} \bigl(1 - S(i,k)\bigr),
\qquad j<i.
\]
The weight assigned to position $j$ is therefore reduced whenever a later position $k$ with $j<k<i$ has a high score.  
As a result, attention concentrates on the rightmost high-scoring predecessor of $i$.  
When the scores are $0$ or $1$, stick-breaking attention behaves like masked hard attention,
selecting the most recent position with score $1$~\cite{tan2025scaling}.
%effectively selecting a single most recent position~\cite{tan2025scaling}.
%
In the two architectures considered in this paper, attention is always strictly masked ($j<i$) and implemented using stick-breaking attention. For comparison, we also evaluate standard softmax-based attention mechanisms.

%% file: task.tex
% Vicenc: I have removed comments (old file is task_old.tex)
\label{sec:task}
The learning task is to uncover the atoms $F$ and the structure of the
actions $A$ (preconditions and effects) from a set $T$ of
\emph{positive} and \emph{negative} action traces over $A$ drawn
from instances $P=\tup{F,A,I,G}$ of a hidden \strips model $M=\tup{F,A}$.
A trace over $A$ is an action sequence $\trace$
made up of actions from $A$.
For convenience and without loss of generality, we ignore the initial situation $I$ and the goals $G$
and define traces to be positive or negative according to their internal consistency,
defined as follows:
\begin{definition}[Positive and negative traces]
Given a propositional \strips model $M=\tup{F,A}$, a trace $\tau=(a_0,\ldots,a_{n-1})$ is \emph{positive}
% \footnote{\VG{justify here why we use of $0$ to denote a positive trace without being too technical. Maybe consider using "consistent" instead of "positive"?}}
if for every atom $p \in \mathrm{pre}(a_i)$, $i=1, \ldots,n$, the following holds:
\begin{enumerate}
\item the last action $a_j$ preceding $a_i$ in $\tau$ that affects $p$ makes $p$ true; i.e., $p \in \mathrm{add}(a_j)$, or
\item no preceding action in $\tau$ affects $p$, i.e., there is no $j<i$ such that $p \in \mathrm{add}(a_j)\cup \mathrm{del}(a_j)$.
\end{enumerate}
The action trace $\tau$ is negative in $M$ if it is not positive. 
\end{definition}

From the perspective of next token prediction, if we assume tokens to be the actions in $M$,
a trace $\tau=(a_1,\ldots,a_{n})$ is positive iff each action $a_i$ is a \emph{possible next action} (applicable) given
the previous actions in the trace, and negative otherwise. We also say that a negative trace $\tau$ is \emph{inconsistent} with the model $M$,
and capture this in  the function $f_M(\tau)$ that is $1$ if $\tau$ is inconsistent, and else is $0$. 

%A positive action trace in a \strips model $M=\tup{F,A}$ is an applicable action sequence in any instance $P=\tup{F,A,I,G}$ of the model where in the initial (dummy) situation $I$ no atom is true.

\begin{definition}[Learning task]
Given a set of positive and negative traces $T$ drawn from a hidden domain $M=\tup{F,A}$,
the task is to learn a Boolean function $f$  over all action sequences $\tau$ over $A$
such that  $f(\tau)=f_M(\tau)$. 
\end{definition}

Learning a  function equal  to $f_M(\tau)$ is equivalent to solving the next token prediction problem in the propositional  \strips setting
where actions are the tokens.  Indeed,  $b \in A$ is a \emph{possible next action} after a positive trace $\tau$ in~$M$ iff  $f_M(\tup{\tau,b})=0$, 
where $\tup{\tau,b}$ is the trace $\tau$ extended with action $b$. 
In the next two sections, we introduce two learning architectures that represent
$f(\tau)$ as a  function $f_\theta(\tau)$ with parameters $\theta$ that will be learned
in a supervised manner from positive and negative traces. 
% Carlos: removed line below to save up space
%We describe these architectures first, and then how they are trained.

% We will show that they can learn propositional \strips models from traces alone.

%Given a trace $\tau=(a_0,\ldots,a_n)$, we write $g_\theta(\tau,i)\in[0,1]$ for the predicted inapplicability of action $a_i$ after prefix $\tau_{\le i-1}$.
%The trace-level prediction $f_\theta(\tau)$ is obtained by requiring all actions to be applicable:
%\[
%f_\theta(\tau)=\prod_{i=0}^{n}\bigl(1-g_\theta(\tau,i)\bigr).
%\]

%% file: architecture.tex
\label{sec:strips_trans}

%Our first proposed architecture builds on the connection between hard-attention transformers and the B-RASP language~\cite{b-rasp}. We call this architecture the \strips Transformer (see Supplementary Material). 
% Carlos: better also make explicit STRIPS here
Our first proposed architecture builds on the connection between hard-attention transformers, the B-RASP language~\cite{b-rasp} and \strips (for more information,
refer to the Appendix).
We call this architecture the \strips Transformer. 
Indeed, checking whether a trace is positive reduces to identifying, for each action and each of its preconditions, the most recent preceding action that affects the corresponding atom, since only that action determines its current truth value. This recency-based dependency aligns naturally with masked hard attention. 
%where each position can attend selectively to earlier positions. 
In the \strips Transformer, attention retrieves, for each atom and trace position, the rightmost prior action that modifies that atom and uses its effect to determine applicability.

We proceed constructively. Starting from an arbitrary propositional \strips model $M=\tup{F,A}$, we describe the set of computations that map a trace $\tau$ to the Boolean value $f_M(\tau)$, and show how these computations are realized as a parametrized transformer with parameters $\theta=\theta_M$ (to be defined next). Each step corresponds to a real-valued counterpart of a Boolean operation in B-RASP. The complete B-RASP program from which the architecture is derived is given in Section~\ref{sec:brasp_program} of the Appendix.

Let $l = 1,\ldots,|F|$ index the atoms and $m = 1,\ldots,|A|$ index the actions in $M=\tup{F,A}$.
The \strips Transformer is a single-layer, multi-head architecture that applies one attention head $l$ per domain atom $p_l$.
We define a real-valued tensor $\theta_M~\in [0,1]^{|F| \times |A| \times 3}$ to encode $M$, where
each parameter $\theta_M(l,m,j)$ encodes whether atom $p_l$, associated with head $l$, is a precondition,
a positive effect, or a negative effect ($j=1,2,3$) of action $a_m \in A$. Formally,

\begin{align}
\theta_M(l, m, 1) &:= 
\begin{cases}
1 & \text{if } p_l \in \mathrm{pre}(a_m) \\
0 & \text{otherwise}
\end{cases} \label{eq:thq} \\
\theta_M(l, m, 2) &:= 
\begin{cases}
1 & \text{if } p_l \in \mathrm{add}(a_m) \cup \mathrm{del}(a_m) \\
0 & \text{otherwise}
\end{cases} \label{eq:thk} \\
\theta_M(l, m, 3) &:= 
\begin{cases}
 1 & \text{if } p_l \in \mathrm{del}(a_m) \\
% 1 & \text{if } p_l \in \mathrm{add}(a_m) \\
0 & \text{otherwise}
\end{cases} \label{eq:thv}
\end{align}

% Each parameter represents the  analogue of the logical conditions defined
% for queries~\eqref{eq:thq}, keys~\eqref{eq:thk} and values~\eqref{eq:thv} in 
% the \brasp formulation above.

%The \strips model $M$, however, is not known, and the tensor of $\theta$ parameters cannot be set from $M$ but have to be learned from positive and negative traces drawn from $M$.

% Note that these parameter values correspond to the optimal configuration 
% $\theta^*$, but can only be defined when the  model~$M$ is known.
% In the next section, we will describe how to learn  $\theta$ from the traces when $M$ is unknown.

%Note that we need access to $M$ to define these parameters.
%In the next section we will describe how to optimize them when $M$ is unknown.

%We first introduce the \strips Transformer using real-valued parameters 
%$\theta$. For clarity, we will initially assume that the number of propositions $|P|$ and actions $|A|$ is 
%known. 

%\subsection{Architecture}

Let $\tau=(a_0,\dots,a_{n-1})$ denote the input trace. Each head~$l$ operates as follows:
%	, treating each proposition independently.  
%
\Omit{ Hector: I don't see what line below, commented, adds 
  Given a trace $\trace$, each head $\text{att}_{p_l}$ produces a vector  
\begin{align}\label{eq:layer}
(y_1, \ldots, y_n)_{p_l} & = \text{att}_{p_l}(a_1, \ldots, a_n),
\end{align}  
and the final output $f_\theta(\tau)$ is obtained by combining the outputs of all heads.\hector{** No mention of attention
  heads in B-RASP section; see this **}.
}
%
%
%The \strips Transformer applies one attention head per domain proposition.  
%Given a trace $\tau = (a_1, \ldots, a_n)$, each head $\text{att}_{p_l}$ produces a vector  
%$$(y_1, \ldots, y_n)_{p_l} = \text{att}_{p_l}(a_1, \ldots, a_n),$$  
%and the final output $f_\theta(\tau)$ is obtained by combining these per-head outputs.  
%
%Each head  is associated with a proposition $p_l$ indexed by $l$, and 
%operates as follows:
\begin{itemize}
    \item \textbf{QKV projection.}  
    For each action $a_i \in \tau$, let $m$ be its index in~$\theta_M$.
    The vectors
    $Q_{p_l}, K_{p_l}, V_{p_l} \in [0,1]^n$ are computed as:
    \begin{align}
        Q_{p_l}(i) &:= \theta_M(l, m, 1), \label{eq:transQ}\\
        K_{p_l}(i) &:= \theta_M(l, m, 2), \label{eq:transK}\\
        V_{p_l}(i) &:= \theta_M(l, m, 3), \label{eq:transV}
    \end{align}
    %\footnote{\hector{** Two observations. 1. Display all equations/assignmnets  with their equation numbers; right now this done only for $y(i)$, etc, but some display with no numbers, and some not even displayed. 2. I'd write Eqs~\ref{eq:rasp_query}--\ref{eq:rasp_value} or equations \eqref{eq:rasp_query}--\eqref{eq:rasp_value}, but I wouldn't enclose equation numbers in parentheses when preceded by Eq. or Eqs. **}}
    
    \item \textbf{Attention Scores.}  
    The attention scores are computed as
    \begin{align}\label{eq:tr_scores}
    %S_{p_l} & := \mathbf{1}\mathbf{1}^\top - Q_{p_l}\,(\mathbf{1}-K_{p_l})^\top,
    S_{p_l} & := Q_{p_l}\cdot K_{p_l}^\top,
    \end{align}
    resulting in a matrix $S_{p_l} \in [0,1]^{n \times n}$, 
    where $S_{p_l}(i,j)=1$ if $p_l$ is a precondition of $a_i$ and $a_j$ affects $p_l$, and $S_{p_l}(i,j)=0$ otherwise.
    %with each entry $S_{p_l}(i,j)$ nonzero exactly when $p_l$ is a precondition of $a_i$ and $a_j$ affects $p_l$.    
    % quantifying how much action $a_j$ affects $a_i$ regarding atom $p_l$.
    % previous implementation
    \Omit{
        \item The attention scores are computed as
        \begin{align}\label{eq:tr_scores}
        %S_{p_l} & := Q_{p_l}\cdot K_{p_l}^\top,
        \end{align}
        resulting in a matrix $S_{p_l} \in [0,1]^{n \times n}$, with each entry
        $S_{p_l}(i,j)$ quantifying how much action~$a_j$ affects action $a_i$ regarding
        atom $p_l$. This corresponds to $Q_p(i) \land K_p(j)$ in Eq.~\ref{eq:rasp_attention}.
         % In other words, $S(i,j)$ represents how likely it is that $\hat{p} \in \mathrm{pre}(a_i)$ (i.e., $Q(i)$) \textit{and} $\hat{p} \in \mathrm{add}(a_j) \cup \mathrm{del}(a_j)$ (i.e., $K(j)$). We know that $a_j$ can only affect $a_i$ if it appears before in the trace (i.e., if $j < i$). Therefore, we apply strict future masking, by setting $S(i,j)=0$ if $j \geq i$.
    }
    \item \textbf{Strict future masking.}
    We apply strict future masking by setting $S_{p_l}(i,j) := 0$ if $j \geq i$, which encodes that actions $a_i$ can only be affected by preceding actions~$a_j$ ($j < i$). % This corresponds to the $j < i$ mask in Eq.~\ref{eq:rasp_attention}.

    \item \textbf{Stick-breaking attention.}  
    To implement hard-attention with scalar scores, we use stick-breaking attention~\cite{tan2025scaling}.  
    Given $S_{p_l}$, this computes:
    \begin{align}\label{eq:stick-breaking}
    S'_{p_l}(i,j) := S_{p_l}(i,j) \cdot \prod_{k=j+1}^{i-1} \bigl(1 - S_{p_l}(i,k)\bigr).
    \end{align}
    This ensures that attention focuses on the rightmost high-scoring position preceding $a_i$.
    %, in agreement with the  $\blacktriangleright_j$ operator in \brasp used in Eq.~\ref{eq:rasp_attention}.
    Note that no explicit $\max$ operation is required.
    %This type of attention will be also analyzed in the Vanilla Transformer later in Section~\ref{sec:vanilla}.

%    Assume that $a_i$ contains $\hat{p}$ as a precondition, and two previous actions $a_j, a_k$ ($k < j < i$) contain $\hat{p}$ as an effect. Then, when $a_i$ is executed, the truth value of $\hat{p}$ will be given by the sign of the effect of $a_j$ (i.e., $\hat{p}$ will be true if $\hat{p} \in \mathrm{add}(a_j)$ and will be false otherwise) and will \textit{not} depend on the effect of $a_k$. In other words, for a given precondition $\hat{p} \in \mathrm{pre}(a_i)$, $a_i$ \textit{only} needs to ``pay attention'' to the most recent action in the trace that affects $\hat{p}$ ($a_j$ in our example). To encode this \textit{recency bias}, we resort to stick-breaking attention \cite{tan2025scaling}. Given the (masked) attention matrix $S$, this method obtains a normalized attention matrix $S'$ according to the following formula:
\item \textbf{Value aggregation.}  
The output vector for head $p_l$, denoted $(y_0, \ldots, y_{n-1})_{p_l}\in [0,1]^n$, is
computed as the matrix-vector product 
\begin{align}\label{eq:str_val}
    y_{p_l} := S'_{p_l} \cdot V_{p_l}.
\end{align}

Each entry $y_{p_l}(i)$ indicates the degree to which action~$a_i$ in the input sequence is \textit{inapplicable} relative to the atom~$p_l$. This will be the case if $p_l$ is a precondition of~$a_i$ 
and the rightmost action preceding $a_i$ deleted $p_l$.
% 
% This corresponds to $Y_p$ in Eq.~\ref{eq:rasp_attention}.
%    where $y_i^{(p_l)}$ represents the likelihood that $a_i$ is inapplicable with respect to $p_l$.
%    \item \textbf{Value aggregation.} Finally, we compute $Y = S'V$ to obtain the final output $(y_1, \ldots, y_n)_{\hat{p}_l}$ of the $l-$th attention head, where $(y_i)_{\hat{p}_l}$ represents the likelihood of $a_i$ being inapplicable when only considering the proposition $\hat{p}_l$.
\end{itemize}
%Therefore, the combination of queries, keys and values encoded in $\theta$ define the preconditions and effects (including sign) for all actions $b \in A$ and propositions $\hat{p} \in \hat{P}$. 

Head outputs are combined via a real-valued OR, indicating for each position~$i$ the degree to which action~$a_i$ is \emph{inapplicable} given its predecessors.
We make explicit the dependence on parameters~$\theta$:
%The outputs of all heads are then combined into a single vector~$y$ using a  real-valued analogue of the logical OR. This step %equivalent to Eq.~\ref{eq:rasp_or_atoms}, 
%indicates for each position $i$ the degree to which the action $a_i$ is \emph{inapplicable} given the previous actions in the sequence.
%We make explicit the dependence on the parameters $\theta$:
\begin{align}\label{eq:transf_output}
 % y(i)&:= 1-\exp\!\left(\sum_{l=1}^{|F|}\log\!\big(1-y_{p_l}(i)\big)\right)\\
 y_\theta(i) & := 1 - \prod_{l=1}^{|F|} \bigl(1 - y_{p_l}(i)\bigr).
% y(i) & :=  \prod_{l=1}^{|F|} \, y_{p_l}(i)  
\end{align}

The final step aggregates the inapplicability degrees $y(i)$ via a real-valued OR. The result is $0$ if and only if every action is applicable:
\begin{align}\label{eq:ftheta}
f_\theta(\tau) := 1-\prod_{i=0}^{n-1} \left(1 - y_\theta(i)\right).
\end{align}

While $y_\theta(i)$ denotes the inapplicability of the action at
position $i$, $f_\theta(\tau)$ aggregates these to represent the probability that
$\tau$ is a negative trace. For $\theta=\theta_M$, $f_\theta(\tau) = f_M(\tau)$, meaning the model $f_\theta$ classifies traces in perfect agreement with $M$:

%While $y_\theta(i)$ above captures the probability that the $i$-th action in the trace $\tau$ is not applicable,
%$f_\theta(\tau)$ captures the probability that some action is the trace is not applicable, and hence, the probability
%that $\tau$ is a negative trace. For the parameters $\theta=\theta_M$, it is easy to show that $f_\theta(\tau)=f_M(\tau)$,
%meaning that $f_\theta$ classifies the traces in agreement with the model $M$:

% The \strips Transformer, as layed out above, represents the \strips model $M$, that is, for every positive trace $\tau$ from $M$, we have $f_\theta(\tau) = 0$, and for every negative trace $\tau$ from $M$, we have $f_\theta(\tau) = 1$.

\begin{theorem}%[From \strips to Transformer]
  Let $M=~\langle F, A \rangle$ be a propositional \strips model.
  Then for $\theta=\theta_M$, as  in Eqs.~\ref{eq:thq}--\ref{eq:thv},
%   and let  $\theta^* \in \{0,1\}^{|F| \times |A| \times 3}$  be \tau$,
$f_{\theta}(\tau) = f_M(\tau)$ for any trace $\tau$ drawn from $M$.
\end{theorem}

% Note that this also implies that for a given prefix $\tau_{\le i-1}$.

% For each position $i$, the model prediction is $y_\theta(\tau_{\le i})\in[0,1]$, expresses the 
% probability that $a_i$ is inapplicable. \hector{Introduce this notation; is $y$ the one above? Say what $\tau_{\leq i}$ is, even if intuitive}.

% \hector{Does  the loss use $y_\theta$ or $f_\theta$? Also, if it's $y$ then use above $y_\theta$. But then why $f_\theta$? Also, $y(i)$

% Carlos: summarized previous paragraph (see below)
Since the operations in Eqs.~\ref{eq:transQ}--\ref{eq:ftheta} are end-to-end differentiable,
the same architecture will allow us  to learn $M$ (encoded as $\theta_M$) from positive and negative traces drawn from $M$.
These traces contain actions from  $A$ but  no information  about the  states. 
This learning process is described in Section \ref{sec:learning}.

  \medskip

%% file: sb_transformer.tex
\label{sec:sb_transformer}
Our second proposed architecture is a standard decoder-style transformer~\cite{transformers}
with one difference: we replace the standard softmax attention and positional encodings
with stick-breaking attention~\cite{tan2025scaling}.
%under strict causal masking.
We refer to this model as the Stick-Breaking (SB) Transformer.

In contrast to the \strips Transformer,
the SB Transformer does not embed a \strips model directly in its architecture, 
%the SB Transformer contains no built-in structural parameters encoding a \strips model.
it does not align attention heads with domain atoms,
and it  does not represent action preconditions or effects explicitly in
the parameters. As we show later, however, a \strips model can be extracted
from a trained SB Transformer, provided that that  the set of actions
is extended with auxiliary actions that encode initial and final states,
in a suitable way.
%
%Both the SB Transformer and the \strips Transformer perform the same type of mapping: given a trace~$\trace$ they output a scalar $f_\theta(\tau)$ that predicts the (in)applicability of all actions in the trace.
%
The SB Transformer consists of:
\begin{itemize}[leftmargin=2.5em]
\item[(i)] a learned token (action) embedding matrix of dimension $d_{\text{model}}$. This replaces the structured query, key, and value tensors used in the \strips Transformer to encode preconditions and effects (Eqs.~\ref{eq:transQ}--\ref{eq:transV});
\item[(ii)] $L$ stacked transformer blocks with $H$ attention heads per block. Each head implements stick-breaking self-attention with strict future masking, followed by additional transformations. In contrast, the \strips Transformer uses a single attention layer ($L=1$) with one head per atom in $F$; and
\item[(iii)] a final linear layer with a sigmoid activation function that replaces the real-valued OR aggregation used in the \strips Transformer
(\ref{eq:transf_output}). 
\end{itemize}

%given a nonempty prefix $\tau_{\le i}$ of a trace
%$\tau=(a_0,\ldots,a_{n-1})$, they output a scalar
%$y_\theta(\tau_{\le i})\in[0,1]$ interpreted as the probability that the last action $a_i$
%is inapplicable given the preceding actions.

More precisely, given a trace~$\trace$,
the SB Transformer produces ``contextual representations''
(embeddings) $h_i^{(\ell)} \in \mathbb{R}^{d_{\text{model}}}$ 
for each position $i=0,\ldots,n-1$ and layer $\ell=0,\ldots,L$.

At layer $\ell=0$, each token $a_i$ is mapped to its learned embedding,
yielding representations $h_i^{(0)}$.
For $\ell=1,\ldots,L$, the representations are updated by

\[
h_0^{(\ell)},\ldots,h_{n-1}^{(\ell)}
=
\mathrm{Block}^{(\ell)}
\bigl(h_0^{(\ell-1)},\ldots,h_{n-1}^{(\ell-1)}\bigr),
\]

\noindent where each block consists of multi-head
self-attention using stick-breaking normalization and strict future masking,
followed by a position-wise feed-forward network (MLP) %  of width $d_{\text{ff}}$
with GELU activation \cite{hendrycks2016gaussian}.
Layer normalization (pre-norm) and residual connections
are applied in the standard way.

After the final layer, each position $i$ has an embedding 
 $h_i^{(L)}$ that depends only on the prefix $\tau_{\le i}=(a_0,\ldots,a_i)$.
%A learned linear readout vector $w$ and bias $b$ produce
Then, a linear layer with learnable parameters $w$ and $b$ yields

\[
y_\theta(i)
=
\sigma\!\bigl(w^\top h_i^{(L)} + b\bigr)
\in [0,1],
\]

\noindent where $\sigma$ is  the sigmoid function.
The value $y_\theta(i)$ represents the probability of action $a_i$ being inapplicable given the previous
actions $a_j$ ($j<i$) in the trace $\tau$.
%This quantity is interpreted as the predicted probability that action $a_i$ is inapplicable given its prefix.

The trace-level prediction is defined exactly as in Eq.~\ref{eq:ftheta}:

\[
f_\theta(\tau)
=
1 - \prod_{i=0}^{n-1} \bigl(1 - y_\theta(i)\bigr),
\]

\noindent where $f_\theta(\tau)$ aggregates the inapplicability degree $y_\theta(i)$ of each action $a_i \in \tau$ via a real-valued OR to obtain the probability that $\tau$ is a negative trace.

%% file: learning.tex
\section{Learning}\label{sec:learning}

\label{sec:loss}

We train the \strips and SB transformers in a supervised manner by comparing the transformer predictions $y_\theta(i)$ with the given labels $z_i\in\{0,1\}$, where $z_i=0$ if action $a_i$ is applicable according to the hidden, ground-truth model and $z_i=1$ otherwise.
%We train the \strips and SB transformers in a supervised manner
%by comparing the model predictions $y_\theta(i)$, that
%represent the probability that action $a_i$ is inapplicable
%in the model, with the given ground-truth labels $z_i\in\{0,1\}$.
%
% Recall that $y_\theta(\tau)\in[0,1]$
% stands for  the probability that the last action in the trace  $\tau$
% is inapplicable.
% During training, we apply the model to full traces under
% causal masking, so that predictions for all positions are computed in
% parallel, as in standard transformer training.
% \subsection{Learning Objective}
%
We use the focal loss \cite{lin2017focal}, a variant of the well-known binary cross-entropy loss:
% over all labeled actions:

\begin{equation}\label{eq:learning-loss}
\mathcal{L}(\theta)
=
\frac{1}{|T|}
\sum_{\tau\in T}
\frac{1}{|\tau|}
\sum_{i=0}^{|\tau|-1}
%w_i\, %w_i is now removed, since we talk about test-p actions later
\ell\!\left(y_\theta(i), z_i\right),
\end{equation}
%where $|\tau|=n+1$ and 

\noindent where the normalization by $|\tau|$ ensures that
all traces contribute equally regardless of length. The focal loss term
$\ell(y,z)$ is:
\begin{equation}\label{eq:focal-bce}
-\alpha\, z(1-y)^{\gamma}\log y
-(1-\alpha)(1-z)y^{\gamma}\log(1-y),
\end{equation}

\noindent where parameter $\alpha\in[0,1]$ controls the relative importance assigned
to inapplicable versus applicable actions, while
$\gamma\ge 0$ reduces the contribution of well-classified examples,
focusing learning on harder cases.
In our experiments, we use $\gamma=1$ and $\alpha=0.999$.
% Carlos: we do not talk now about w_i since test-p actions are explained latter.
%The weights $w_i\ge 0$ allow different types of actions
%(e.g., domain actions and \emph{test-p} actions)
%to contribute proportionally to the loss.
% In particular, they are used to control the relative influence
% of domain and goal-test actions during training.

During training, we apply an additional masking rule:
if $z_j=1$, then actions $a_i$ at later positions $i>j$ cannot attend to $a_j$ when performing self-attention,
reflecting the fact that inapplicable actions do not change the state.
%then for any later position $i>j$, the token
%at position $j$ is masked out of the self-attention of
%position $i$,
As this requires ground-truth labels, it is done only during training.

%% file: planning.tex
We now describe the extraction of a \strips\ model $M'=\tup{F',A'}$ from the trained transformers for its use in planning.
A key challenge is that planning problems $P=\tup{F,A,I,G}$ are defined in terms of the  set of atoms  in $F$ in the hidden
ground-truth model $M=\tup{F,A}$, whereas  the traces used in training  convey information about  $A$ but not $F$. 
For solving planning problems $P$ over the hidden model $M$ it is thus necessary to align the two models $M$ and $M'$,
so that problems $P$ over $M$ can be mapped into equivalent problems $P'$ over the learned model $M'$.
For this, we  augment the set of actions $A$ used during  training with auxiliary \textit{setup actions}
that convey information about the initial and final states of the traces. No information
is conveyed about intermediate states. 

%We describe now how a \strips model $M'=\tup{F',A'}$ can be extracted from the trained \strips and SB transformers, to then use for planning.
%However, we first need to address a crucial point.
%Planning problems $P=\tup{F,A,I,G}$ are encoded in terms of atoms $p \in F$ of the hidden, ground-truth model $M=\tup{F,A}$, but training traces only contain information %about the actions $A$.
%To address this gap, we need to also incorporate information about $F$ in the training traces.
%We do so by augmenting training traces with special \textit{setup actions}, different to the domain actions in $A$, as we explain below.

\iffalse
Planning with the learned models is not direct. We have the
hidden model $M=\tup{F,A}$, and want to  a plan for a
hidden instance $P=\tup{F,A,I,G}$ without having the model $M$
but a learned model $M'=\tup{F',A'}$ instead. The two models share
the set of actions $A=A'$,  but not  the set of atoms.
We address next the problem of specifying
a problem $P'=\tup{F',A',I',G'}$ over the learned model $M'$
such that the plans obtained from $P'$ with an off-the-shelf
planner can be validate on the hidden instance $P$.
\fi

\subsection{Initial and Final States in the Traces}

% To incorporate state information into a training trace $\tau=(a_0, \dots, a_{n-1})$, we augment the domain actions $A$ with special setup actions.

There are three types of auxiliary (setup) actions: \textit{init-false}, \textit{init-p}, and \textit{test-p} for $p \in F$.  The first action
sets all atoms $p$ in $F$ to false, the second has $p$ as the single effect, and the third has no effects. The first two actions, in turn, have no preconditions,
while  \textit{test-p}  has precondition $p$. The action traces used in training,  which are drawn
from an initial state $s_0$ and end in a final state $s_n$,  are extended with a \emph{prefix} that conveys information about $s_0$,
and a \emph{suffix} that conveys information about $s_n$. More precisely, the  prefix contains the action \textit{init-false}
followed by actions \textit{init-p} for each $p$ true in $s_0$, while the suffix contains all actions \textit{test-p} for $p \in F$,
in the case of the SB transformer, and all actions \textit{test-p} for $p \in G$, where $G$ is the goal of the instance,
in the \strips transformer. The actions \textit{test-p} have precondition $p$ and hence, they  are labeled as  applicable in the state $s_n$ and hence
in the training trace iff $p$ is true in $s_n$. The transformers are trained to predict the applicability of both the domain and
the setup actions, and thus they must   learn  to predict the truth values of all or some atoms at the end of each trace,
as they determine the applicability of the \textit{test-p} actions.

In the \strips\ transformer, the model of the auxiliary actions, i.e., their preconditions and effects, can be hardwired and does not have to be learned.
This is achieved by  mapping the atoms $p \in F$ to different transformer heads, and  setting the value of their $\theta$ parameters accordingly. 
In the SB transformer, this is not possible and the model of the auxiliary actions has to be learned like the model of the domain actions,
as there is no one-to-one correspondence between model parameters and symbolic preconditions or effects.

\subsection{Extracting the \strips Models}
\label{sec:model_extraction}

In the \strips Transformer, the model $M'=\tup{F',A'}$ is extracted from the trained parameters $\theta$ as follows. Let $l = 1,\ldots,|F'|$ index the attention heads, and $m = 1,\ldots,|A'|$ index the actions.
We binarize the tensor of parameters $\theta$ as

%In the \strips transformer the action preconditions and effects are extracted from the training model as follows.
%Let $l = 1,\ldots,|F|$ index the number of heads of the \strips transformer and $m = 1,\ldots,|A|$ index the actions observed in the data.

\begin{align}\label{eq:binarized}
{\bar{\theta}}(l, m, k) = \llbracket \theta(l, m, k) \geq 0.5 \rrbracket,
~\quad\text{for $k = 1, 2, 3$},
\end{align}

\noindent and set the preconditions and effects of actions $a_m \in A'$ as
% \begin{definition} (From \strips Transformer to \strips model).
% The \strips model $M_{\bar{\theta}}$ learned by the \strips Transformer
% has as many atoms as heads and as many actions as observed in the data.
% 
\begin{align*}
    \mathrm{pre}({a}_m) & = \{{p}_l \mid \bar{\theta}(l,m,1)\},\\
     \mathrm{add}({a}_m) & = \{{p}_l \mid \bar{\theta}(l,m,2) \land \neg\bar{\theta}(l,m,3)\},\\
    \mathrm{del}({a}_m) & = \{{p}_l  \mid \bar{\theta}(l,m,2) \land \bar{\theta}(l,m,3)\}.
  \end{align*}
    \label{def:Mb}

\Omit{
 To extract a symbolic model from a trained SB Transformer, we use a
\emph{state-probing construction} that reconstructs the sequence of states associated with a trace from predicted individual actions applicability. Since the SB Transformer does not provide direct access to the internal structure of the state, the training traces for the SB Transformer need to be more informative, containing a set~\emph{end-setup} actions consisting of one $\textit{test-}p$ action  with precondition~$\{p\}$ and no effects for every atom $p \in F$.
}

\medskip

For the SB Transformer, the \strips model is extracted in two stages. First, a computationally intensive \textit{state probing} process is performed, wherein the model predicts the applicability of each \textit{test-p} action for every prefix $\tau_{\le t}$ of the training traces.
%
%For the SB transformer, the \strips model is extracted in two steps. 
%First, a computationally expensive \textit{state probing} process is required, where the transformer predicts the applicability of each \textit{test-p} action for every prefix $\tau_{\le t}$ of the training traces.
%This is equivalent to predicting the truth value of every atom $p \in F$ in the state $s_i$ after executing $\tau_i$.
% 
More precisely, if $\tau = (a^I_0, \dots, a^I_r, a_0, \dots, a_{n-1})$ is a training trace containing the  initial setup actions $(a^I_0, \dots, a^I_r)$ but no end setup actions,the reconstructed state $s_t^\tau$ at time $t\in\{-1,0,\dots,n-1\}$ is defined as:

\[
s_t^\tau = \{\, p \in F \mid y_\theta(\tup{\tau_{\le t}, \textit{test-}p}) < 0.5 \,\},
\]

\noindent where $\tau_{\le t}=(a^I_0, \dots, a^I_r, a_0,\dots, a_t)$. The model $M'=\tup{F,A}$ is then derived from these reconstructed states in a standard way, as follows.

Let $\mathrm{eff}(p,s,s')$ denote a function that returns $\mathrm{add}$ if $p \notin s \wedge p \in s'$, $\mathrm{del}$ if $p \in s \wedge p \notin s'$, and $\mathrm{none}$ otherwise. For each atom $p \in F$ and action $a \in A$, the associated preconditions and effects are defined by majority consensus over the training data: 
\begin{itemize}
    \item $p \in \mathrm{pre}(a)$ iff $p \in s_{t-1}^\tau$ for at least $95\%$ of transitions where $a_t=a$;
    \item $p \in \mathrm{add}(a)$ (resp. $p \in \mathrm{del}(a)$) iff $\mathrm{add}$ (resp. $\mathrm{del}$) is the most frequent output of $\mathrm{eff}(p,s_{t-1}^\tau,s_{t}^\tau)$ for $a_t=a$.
\end{itemize}

For both the \strips and SB transformers, the model extraction process described here results in \strips planning domains such as those shown
in Listing \ref{lst:pddl_domains}.

\subsection{Planning with the Learned Model}

Once the \strips model $M'=\tup{F',A'}$ is extracted, it can be used to solve planning problems $P=\tup{F,A,I,G}$ over the hidden
domain $M=\tup{F,A}$ by defining an equivalent problem $P'=\tup{F',A,I',G'}$, which like $P$ does not include the auxiliary actions.
For the SB Transformer, $F'$ is $F$, as the former is obtained from the set of \textit{test-p} actions used in training for all $p \in F$.
As a result,  $I'$ and $G'$ can be set in $P'$ to $I$ and $G$, respectively.
In the \strips Transformer, on the other hand,  we include in $I'$ the learned atom $p' \in F'$ corresponding to the single (hardwired) add effect of the
action \textit{init-p}, for each $p \in I$, and in $G'$, the learned atom $p' \in F'$ corresponding to the single (hardwired)  precondition of the action \textit{test-p}.

% Given a correctly learned model $M'$, an off-the-shelf planner can solve $P'$, yielding a plan that is also a valid solution for the original problem $P$.

An off-the-shelf classical planner can then be used to solve the problem  $P'$. If the learned model $M'$ is correct, the resulting plan 
will be a plan for the target problem $P$ too. In our experiments, we used the Mimir planner~\cite{mimir} with greedy best-first-search (gbfs) and
the FastForward (FF) heuristic~\cite{hoffmann2001ff}.

%and, unlike for \textit{init-false} and \textit{init-p} actions, these need to be learned by the \strips Transformer.
%We perform experiments with the \textit{init-state} encoding for completeness since, in this setting, the initial state $I$ of each problem must be in $S_I$, so we can only evaluate compositional reasoning over goals.

It is worth noticing that the role of the setup or auxiliary  actions is different in  the two architectures. In the \strips\ Transformer, setup
actions are not required to learn the domain model $M'=\tup{F',A'}$, but to align  $F'$ with $F$, and thus to map the target 
planning problems $P=\tup{F,A,I,G}$ over the hidden model $M=\tup{F,A}$ into planning problems $P=\tup{F',A',I',G'}$ over the learned model.
In the SB Transformer, on the other hand, setup actions are essential to learn the \strips model $M'=\tup{F',A'}$ in the first place.
Without such auxiliary actions, the SB Transformer would just give us a Boolean trace classifier for determining whether an action trace
is positive (applicable) or negative (not applicable). The gap is bridged in this case by learning to predict the applicability of
the \textit{test-p} actions, one for each $p \in F$, and hence, the truth value of $p$ along any of the  traces used in training.
The full symbolic states along the traces are thus inferred, and the \strips model is obtained from the resulting interleaved sequences of
actions and states in a standard way, as detailed in Section~\ref{sec:model_extraction}.

\subsection{Alternative Initial State Encoding}

For completeness, we also evaluate the learned \strips models  using a different encoding
of the initial states in the training traces.
In the previous scheme, called \textit{init-atoms}, an  initial state $s_0$ was encoded by means of a prefix made up of 
\textit{init-false} and \textit{init-p}  actions, one for each $p$ in the initial situation $I$. 
The alternative encoding is simpler and  assumes  a predefined set of possible  initial states $S_I$
which is used in both  training and testing.  In this scheme, called \textit{init-state}, there is  one  setup action \textit{init-s} for
each $s \in S_I$ with  no preconditions, and with effects that  make all   the atoms in $s$ true,  and all other atoms false.
The resulting trace prefixes contain in this case  just one auxiliary action, an   \textit{init-s} action,
and both the \strips and SB Transformers must learn its model,  like the model of all domain actions.
In this encoding of the initial states in the traces, and in contrast with the \textit{init-p} scheme,
there is no generalization to an exponential number of possible initial states,
yet  there is  generalization to an exponential number of possible goals.
In the experiments below, these two encoding schemes are evaluated
experimentally for the two architectures.

%% file: experiments.tex
\begin{table}[h!]
% \begin{table}[ht]
\centering
\small
\begin{tabular}{@{}lrrrr@{}}
\toprule
& \multicolumn{4}{c}{Blocksworld} \\
\cmidrule(lr){2-5}
Model & Train & Test & Plan ($M_S$) & Plan ($M_L$) \\
\midrule
SB & \textbf{1.0} (\textbf{1.0}) & .998 (.999) & \textbf{1.0} (\textbf{1.0}) & \textbf{1.0} (\textbf{1.0}) \\
Sinusoidal & .994 (.998) & .237 (.250) & \textbf{1.0} (\textbf{1.0}) & .00 (.00) \\
RoPE & .998 (.999) & .363 (.377) & \textbf{1.0} (\textbf{1.0}) & .00 (.00) \\
\bottomrule
\end{tabular}

\begin{tabular}{@{}lrrrr@{}}
\toprule
& \multicolumn{4}{c}{Ferry} \\
\cmidrule(lr){2-5}
Model & Train & Test & Plan ($M_S$) & Plan ($M_L$) \\
\midrule
SB & \textbf{1.0} (\textbf{1.0}) & .999 (.998) & \textbf{1.0} (\textbf{1.0}) & \textbf{1.0} (\textbf{1.0}) \\
Sinusoidal & .999 (.999) & .258 (.258) & \textbf{1.0} (\textbf{1.0}) & .00 (.00) \\
RoPE & \textbf{1.0} (\textbf{1.0}) & .530 (.541) & \textbf{1.0} (\textbf{1.0}) & .25 (.04) \\
\bottomrule
\end{tabular}

\begin{tabular}{@{}lrrrr@{}}
\toprule
& \multicolumn{4}{c}{Npuzzle} \\
\cmidrule(lr){2-5}
Model & Train & Test & Plan ($M_S$) & Plan ($M_L$) \\
\midrule
SB & \textbf{1.0} (\textbf{1.0}) & .997 (.997) & \textbf{1.0} (\textbf{1.0}) & \textbf{1.0} (\textbf{1.0}) \\
Sinusoidal & .999 (.999) & .253 (.253) & \textbf{1.0} (\textbf{1.0}) & .01 (.01) \\
RoPE & .998 (.999) & .402 (.445) & \textbf{1.0} (\textbf{1.0}) & .02 (.03) \\
\bottomrule
\end{tabular}

\begin{tabular}{@{}lrrrr@{}}
\toprule
& \multicolumn{4}{c}{Maze} \\
\cmidrule(lr){2-5}
Model & Train & Test & Plan ($M_S$) & Plan ($M_L$) \\
\midrule
SB & \textbf{1.0} (\textbf{1.0}) & .958 (.967) & .98 (.98) & .93 (.93) \\
Sinusoidal & \textbf{1.0} (\textbf{1.0}) & .510 (.506)  & .98 (.98) & .00 (.00) \\
RoPE & \textbf{1.0} (\textbf{1.0}) & .578 (.562) & .98 (.98) & .00 (.00) \\
\bottomrule
\end{tabular}

\begin{tabular}{@{}lrrrr@{}}
\toprule
& \multicolumn{4}{c}{Logistics} \\
\cmidrule(lr){2-5}
Model & Train & Test & Plan ($M_S$) & Plan ($M_L$) \\
\midrule
SB & \textbf{1.0} (\textbf{1.0}) & .987 (.993) & \textbf{1.0} (\textbf{1.0}) & \textbf{1.0} (\textbf{1.0}) \\
Sinusoidal & .999 (\textbf{1.0}) & .281 (.284) & \textbf{1.0} (\textbf{1.0}) & .00 (.00) \\
RoPE & .999 (.999) & .412 (.402) & \textbf{1.0} (\textbf{1.0}) & .32 (.06) \\
\bottomrule
\end{tabular}
\caption{Comparison of next-token prediction accuracy and planning accuracy using extracted \strips models.
We report results for each domain using the large problem size and the \textit{init-atoms} encoding.
$M_S$ and $M_L$ denote models extracted from short training traces and longer test traces, respectively. While the SB Transformer achieves near-perfect performance across all metrics, baseline models exhibit poor length generalization in both prediction and symbolic extraction from long traces. Notably, in all three cases, a symbolic model extracted from the trained transformer using short traces ($M_S$) is able to plan with perfect accuracy.}
%Comparison between the SB Transformer and the two baseline models in \bwl with \textit{init-atoms} encoding.
%We provide the training and test accuracy of each model. For the planning accuracy, we evaluate two different \strips models. The first model $M_S$ is extracted by performing state probing on the training traces (with maximum length $D=50$), as in Table \ref{tab:main_results}. The second model $M_L$ is extracted from the test traces, which are longer ($D=200$).
%In both cases, the SB Transformer achieves perfect planning accuracy. On the other hand, the baseline models (\textit{Sinusoidal} and \textit{RoPE}) do not exhibit good length generalization, as reflected by their low test accuracy. As a result, planning accuracy severely degrades when the \strips model is extracted from the longer, test traces.
\label{tab:baselines}
\end{table}

We evaluate the \strips and SB Transformers on trace classification and planning. This section details our experimental setup to address the following research questions:
%We conduct experiments to evaluate the abilities of the \strips and SB Transformers to classify action traces and plan with the learned models. Specifically, we are interested in addressing the following questions:

\begin{itemize}
    \item How do the \strips and SB Transformers compare against standard transformer baselines?
    \item Does the additional built-in structure of the \strips Transformer lead to better results when compared to the SB Transformer?
    \item Do the models exhibit compositional reasoning and generalize to an exponential number of unseen states and goals?
\end{itemize}

To evaluate the performance of our proposed architectures, we compare them against two standard baselines:
\begin{itemize}
    \item \textbf{Sinusoidal:} The original transformer architecture using softmax attention and sinusoidal positional encodings~\cite{transformers}.
    \item \textbf{RoPE:} A transformer using Rotary Position Embeddings~\cite{su2024roformer}. This model encodes relative token dependencies via rotation matrices, which has been shown to enhance generalization on long sequences~\cite{spies2025transformers}.
\end{itemize}

\iffalse
We conduct several experiments to evaluate the capabilities of the \strips and SB Transformers to classify action traces and plan with the learned models. In addition to our models, we consider two baselines:

\begin{itemize}
    \item Standard transformer with softmax attention and sinusoidal positional encodings, as proposed in the original transformer paper~\cite{transformers}.
    \item Standard transformer with softmax attention and Rotary Position Embeddings (RoPE), where the absolute position of each token (i.e., action) in the sequence is encoded through a rotation matrix, allowing self-attention to model relative position dependencies between tokens.
    This has been shown to improve generalization and accuracy on long sequences when compared to other positional encodings~\cite{su2024roformer}.
\end{itemize}

\paragraph{Planning domains.}
We consider five planning domains. For each domain, we define two problem sizes: small instances with $|F|<50$ atoms and large instances with $50 < |F| < 100$ atoms. This allows us to use the same number $H$ of attention heads for the \strips Transformer across domains, setting $H=50$ for small problems and $H=100$ for large problems, ensuring that $H \geq |F|$ for every domain and size. This shows that we do not need to know the number of atoms in the ground-truth domain, provided that $H \geq |F|$. For each size, problems share the same objects and static atoms. Finally, we provide a brief description of each domain below:
\fi

\paragraph{Planning domains.}
We consider five domains, each with two problem sizes: \emph{small} instances ($|F|<50$ atoms) and \emph{large} instances ($50 < |F| < 100$). We set the number of attention heads for the \strips Transformer to $H=50$ and $H=100$ respectively, ensuring $H \geq |F|$ without requiring prior knowledge about the exact number of atoms. Within each size, problems share the same objects and static atoms.
\iffalse
    \begin{itemize}
        \item \textbf{\bw}: this domain involves reassembling a set of stackable blocks into a target configuration with a  gripper. We use two grounded versions: \bws and \bwl, containing five and eight blocks, respectively.
        \item \textbf{\fr}: in this domain a ferry is used to transport cars between different ports. We use two grounded  versions: \frs and \frl, containing five cars and five ports, and eight cars and seven ports, respectively.
        \item \textbf{\np}: a puzzle game in which several numbered tiles must be slid into their goal positions. We use two grounded versions: \nps and \npl, containing five and eight tiles, respectively.
        \item \textbf{\mz}: a grid maze where an agent must traverse between two locations.
        We use a predicate $free$ to encode traversable locations (i.e., which are not walls). We use two grounded versions: \mzs and \mzl, of size 5x5 and 7x7, respectively.
        \item \textbf{\lgt}: logistics task where a fleet of trucks and airplanes is used to deliver packages across locations in different cities. We use two grounded versions: \lgts and \lgtl, containing three cities and two packages, and five cities and four packages, respectively. For both sizes, problems contain a single airplane, and each city contains an airport, a location and a truck.
    \end{itemize}
    \fi
\begin{itemize}
    \item \textbf{\bw}: Reassembling stackable blocks with a gripper. We use two grounded versions: \bws and \bwl, containing five and eight blocks, respectively.
    \item \textbf{\fr}: Transporting cars between ports with a ferry. \frs uses five cars and five ports; \frl uses eight cars and seven ports.
    \item \textbf{\np}: Sliding numbered tiles to goal positions. \nps uses five tiles; \npl uses eight.
    \item \textbf{\mz}: Grid maze traversal with a $free$ predicate for traversable cells. \mzs is $5 \times 5$; \mzl is $7 \times 7$.
    \item \textbf{\lgt}: Package delivery via trucks and airplanes. \lgts contains three cities and two packages; \lgtl contains five cities and four packages.
    For both sizes, problems contain a single airplane, and each city contains an airport, a location and a truck.
\end{itemize}

\paragraph{Data generation.}

For each domain and size, we randomly generate training, test, and planning problems. 
Under \textit{init-state} encoding, initial states are sampled from a predefined set $S_I$.
Under \textit{init-atoms} encoding, problems do not share initial states.
Training and test problems are used to generate corresponding action traces. 
% Already explained in section 7.1
% For each problem, we append to the trace $\tau$ the actions encoding its initial state (\textit{init-sX} or \textit{init-p}, depending on the configuration). Then, we execute $d$ actions at random from the set of propositional actions in the domain, where $d$ is sampled uniformly from $[1,D]$. Lastly, we append to $\tau$ the actions \textit{goal-p} encoding the final state of $\tau$. \textcolor{red}{Explicar antes que Vanilla y STRIPS Transformer usan distintas goal-p actions}
We generate $10^5$ training traces of maximum length $D=50$, and $2\cdot10^5$ test traces with $D=200$ to evaluate length generalization.
For \mzl, we use $D=200$ for training to ensure successful learning, and we test using $D=400$.
Each trace contains $d$ domain actions plus setup actions, with $d$ sampled uniformly from $[0, D]$.
%Given a particular value for $D$, each trace will contain $d$ domain actions in addition to several setup actions, where $d$ is uniformly sampled from the range $[0,D]$.
For training, domain actions are sampled by picking applicable and inapplicable actions with equal probability.
In contrast, in the test set, 
50\% of traces consist only of applicable actions, while the other 50\% contain $d-1$ applicable actions followed by a single inapplicable final action.

% Already explained in Section 7.1
%For the training traces, we generate the domain actions by sampling applicable and inapplicable actions with equal probability in order to avoid class imbalance. \textcolor{red}{Hay que haber explicado que las trazas pueden tener varias acciones inaplicables antes en el paper}
% No sería correcto decir que 50\% of test traces son positivas y el otro negativas, ya que nos referimos explícitametne a las domain actions, pero después de las n domain actions aparecen las goal actions goal-p, algunas de las cuales serán aplicables y otras no aplicables.
% In the test dataset, 50\% of the traces contain only applicable domain actions, whereas the remaining 50\% consist of traces in which the first $d-1$ domain actions are applicable and the final action is inapplicable.
% Técnicamente esta es una forma "práctica" de hacer partial grounding
%Finally, for each domain and size, we filter out those domain actions that are never applicable in any trace, a technique known as \textit{partial grounding}.

\begin{figure*}[t]
    % Make this float behave like a Listing
    \captionsetup{type=listing}
    \caption{\textbf{\strips domains extracted from the \strips and SB transformers.} Listings show the schema for action \texttt{stack\_A\_B} learned by (a) the \strips Transformer and (b) the SB Transformer in \bwl, for the \textit{init-atoms} encoding and seed = 3. Constants \texttt{A} and \texttt{B} represent particular objects of type \texttt{block}. 
    The domain of the \strips Transformer is encoded in terms of atoms \texttt{pX}, different to those of the ground-truth domain. For those \texttt{pX} appearing in initial states, we can obtain their associated ground-truth atom using the \textit{init-p} actions. For those appearing in goals, we can use the \textit{test-p} actions.
    For instance, in (a), the atom \texttt{p2} is translated to \texttt{clear\_B}, whereas \texttt{p80} and \texttt{p99} cannot be translated as they encode learned atoms not present in initial states or goals.
    In contrast, since the SB Transformer requires one \textit{test-p} action for each ground-truth atom $p$,
    the extracted domain uses the same atoms as the ground-truth domain (see (b)).
    %every \texttt{pX} can be translated and, thus, the learned domain uses the same atoms as the ground-truth domain (see (b)).
    %In contrast, the SB Transformer performs state probing on the training traces to extract its domain, requiring the existence of a \textit{test-p} action for each atom $p$ in the ground-truth domain. By comparison, the \strips Transformer requires \textit{test-p} actions only for goal atoms $p$. Consequently, the extracted domain uses the same atoms as the ground-truth domain, so no translation is required (see (b)).
    For the SB Transformer, we can also substitute the objects (\texttt{A}, \texttt{B} in the example) in the learned action schemas by variables (\texttt{?x}, \texttt{?y} in the example) in order to obtain an equivalent set of lifted action schemas (see (c)).}
    \label{lst:pddl_domains}
    \centering
    \begin{minipage}[t]{0.33\textwidth}

\begin{center}
(a) \strips Transformer
\end{center}

\begin{lstlisting}[language=PDDL]
(:action stack_A_B
 :parameters ()
 :precondition (and
     (clear_B)
     (p80)
     (p99))
 :effect (and
     (handempty)
     (on_A_B)
     (not (clear_B))
     (not (p99))))
\end{lstlisting}
    \end{minipage}
    \hfill
    \begin{minipage}[t]{0.33\textwidth}

\begin{center}
(b) SB Transformer
\end{center}
    
\begin{lstlisting}[language=PDDL]
(:action stack_A_B
 :parameters ()
 :precondition (and
  (clear_B)
  (holding_A))
 :effect (and
  (clear_A)
  (handempty)
  (on_A_B)
  (not (clear_B))
  (not (holding_A))))
\end{lstlisting}
    \end{minipage}
    \hfill
    \begin{minipage}[t]{0.33\textwidth}

\begin{center}
(c) SB Transformer (after lifting)
\end{center}
    
\begin{lstlisting}[language=PDDL]
(:action stack
 :parameters (?x ?y)
 :precondition (and
  (clear ?y)
  (holding ?x))
 :effect (and
  (clear ?x)
  (handempty)
  (on ?x ?y)
  (not (clear ?y))
  (not (holding ?x))))
\end{lstlisting}
    \end{minipage}

    \vspace{4pt}
    \hrule height 1.2pt

\end{figure*}

\paragraph{Training and evaluation.}

%We perform $5\cdot10^5$ training steps using the Rectified Adam (RAdam) optimizer \cite{liu2019variance} on batches with 16 traces and 64 traces for small and large problem sizes, respectively. For the \strips Transformer, we use a learning rate of $0.01$ in every domain except $\frl$ with the \textit{init-atoms} encoding, where a smaller value of $0.005$ was required for convergence. For the Vanilla Transformer, we use a learning rate of $5 \cdot 10^{-5}$. 
%We initialize the trainable parameters $\theta$ of the \strips Transformer as follows: values $\theta(:,:,3)$ are sampled from the distribution $U[0,1)$, whereas queries $\theta(:,:,1)$ and keys $\theta(:,:,2)$ are sampled from $U[0,1)$ for the \textit{init-state} encoding and from $U[0,0.1)$ for \textit{init-atoms}. The latter can be seen as initializing actions with (almost) no preconditions or effects. We also apply L1 regularization to $\theta(:,:,1)$ and $\theta(:,:,2)$ with a value of $10^{-4}$.

We train for $5\cdot10^5$ steps using the RAdam optimizer \cite{liu2019variance} with batches of 16 and 64 traces for small and large problem sizes, respectively. For the \strips\ Transformer, we use a learning rate of $0.01$, except for \frl\ with \textit{init-atoms} where we use $5\cdot 10^{-3}$. For the SB Transformer, we use a learning rate of $5 \cdot 10^{-5}$. The parameters $\theta$ of the \strips\ Transformer are initialized as follows: $\theta(:,:,3) \sim U[0,1)$; queries $\theta(:,:,1)$ and keys $\theta(:,:,2)$ are sampled from $U[0,1)$ for \textit{init-state} and $U[0,0.1)$ for \textit{init-atoms}. The latter effectively initializes actions with a minimal number of preconditions and effects. We also apply L1 regularization ($10^{-4}$) to queries and keys.

%Every $2\cdot10^4$ training steps we perform one validation epoch, computing the accuracy of the current model on the entire training dataset. After training, we load the checkpoint with the best training accuracy and calculate its accuracy on the test dataset. A trace $\tau=(a_0, \dots, a_n)$ is correctly classified if the model correctly predicts the applicability for every action $a_{0..n}$, including both domain and end setup actions. Then, accuracy simply corresponds to the percentage of traces that are correctly classified.

Every $2\cdot10^4$ training steps,
we perform a validation epoch by computing model accuracy on the entire training set. After training, we load the checkpoint with the highest training accuracy to evaluate on the test dataset. A trace $\tau=(a_0, \dots, a_{n-1})$ is considered correctly classified only if the model predicts the applicability of \emph{every action} in the trace $a_{0..n-1}$, including domain and setup actions. Accuracy is defined as the percentage of correctly classified traces. 

%Before computing predictions with the \strips Transformer, we first need to obtain the binarized parameters $\bar{\theta}$. In the case of the Vanilla Transformer, we simply round predictions to $\{0,1\}$.
%Lastly, we perform planning with the \strips model extracted from the transformer parameters, obtaining its accuracy on the set of planning problems.
%All experiments were run on a machine equipped with an Nvidia A10 GPU and an Intel Xeon Platinum 8352M CPU.
%Finally, our code will be made public upon paper acceptance.

For the \strips Transformer, predictions are computed using the binarized parameters $\bar{\theta}$, whereas for the SB Transformer predictions are rounded to $\{0,1\}$.
For both architectures, we extract the \strips models  from the learned parameters to evaluate their performance on the planning problems as described in Section~\ref{sec:model_extraction}.
All experiments were conducted on a machine equipped with an Nvidia A10 GPU and an Intel Xeon Platinum 8352M CPU.

%% file: results.tex
\begin{table*}[h!]
\centering
\small
\begin{tabular}{@{}lrrrrrr@{}}
\toprule
 & \multicolumn{6}{c}{Blocksworld} \\
 & \multicolumn{3}{c}{5 blocks} & \multicolumn{3}{c}{8 blocks} \\
\cmidrule(lr){2-4}\cmidrule(lr){5-7}
Model & Train & Test & Plan & Train & Test & Plan \\
\midrule
\strips 50 states & .975 (\textbf{1.0}) & .964 (\textbf{1.0}) & .93 (\textbf{1.0}) & .982 (\textbf{1.0}) & .904 (\textbf{1.0}) & .94 (.98) \\
\strips 100 states & .967 (\textbf{1.0}) & .918 (\textbf{1.0}) & .98 (\textbf{1.0}) & \textbf{1.0} (\textbf{1.0}) & .996 (.996) & .91 (.96) \\
\strips atoms & .615 (.906) & .362 (.864) & .65 (\textbf{1.0}) & .544 (.979) & .286 (.648)  & .53 (.94) \\
\midrule
SB 50 states & .999 (\textbf{1.0}) & .984 (.991) & \textbf{1.0} (\textbf{1.0}) & \textbf{1.0} (\textbf{1.0}) & .982 (.997)  & \textbf{1.0} (\textbf{1.0}) \\
SB 100 states & \textbf{1.0} (\textbf{1.0}) & .985 (.986) & \textbf{1.0} (\textbf{1.0}) & \textbf{1.0} (\textbf{1.0}) & .985 (.994) & \textbf{1.0} (\textbf{1.0}) \\
SB atoms & \textbf{1.0} (\textbf{1.0}) & .995 (.999) & \textbf{1.0} (\textbf{1.0}) & \textbf{1.0} (\textbf{1.0}) & .998 (.999)  & \textbf{1.0} (\textbf{1.0}) \\
\bottomrule
\end{tabular}

\vspace{0.05cm}

\begin{tabular}{@{}lrrrrrr@{}}
\toprule
 & \multicolumn{6}{c}{Ferry} \\
 & \multicolumn{3}{c}{5 cars} & \multicolumn{3}{c}{8 cars} \\
\cmidrule(lr){2-4}\cmidrule(lr){5-7}
Model & Train & Test & Plan & Train & Test & Plan \\
\midrule
\strips 50 states & .985 (\textbf{1.0}) & .901 (\textbf{1.0}) & .95 (\textbf{1.0}) & \textbf{1.0} (\textbf{1.0})  & \textbf{1.0} (\textbf{1.0}) & .98 (.93) \\
\strips 100 states & \textbf{1.0} (\textbf{1.0}) & \textbf{1.0} (\textbf{1.0}) & \textbf{1.0} (\textbf{1.0}) & .992 (\textbf{1.0}) & .930 (\textbf{1.0}) & .88 (.97) \\
\strips atoms & .916 (\textbf{1.0}) & .894 (\textbf{1.0}) & .92 (\textbf{1.0}) & .742 (.821) & .538 (.742) & .28 (.70) \\
\midrule
SB 50 states & .999 (\textbf{1.0}) & .911 (.922) & \textbf{1.0} (\textbf{1.0}) & \textbf{1.0} (\textbf{1.0}) & .950 (.971) & \textbf{1.0} (\textbf{1.0}) \\
SB 100 states & .999 (.999) & .898 (.921) & \textbf{1.0} (\textbf{1.0}) & \textbf{1.0} (\textbf{1.0}) & .968 (.977) & \textbf{1.0} (\textbf{1.0}) \\
SB atoms & \textbf{1.0} (\textbf{1.0}) & .988 (.992) & \textbf{1.0} (\textbf{1.0}) & \textbf{1.0} (\textbf{1.0}) & .999 (.998) & \textbf{1.0} (\textbf{1.0}) \\
\bottomrule
\end{tabular}

\vspace{0.05cm}

\begin{tabular}{@{}lrrrrrr@{}}
\toprule
 & \multicolumn{6}{c}{Npuzzle} \\
 & \multicolumn{3}{c}{5 tiles} & \multicolumn{3}{c}{8 tiles} \\
\cmidrule(lr){2-4}\cmidrule(lr){5-7}
Model & Train & Test & Plan & Train & Test & Plan \\
\midrule
\strips 50 states & .942 (\textbf{1.0}) & .810 (\textbf{1.0}) & .98 (\textbf{1.0}) & .953 (.969) & .810 (.907) & .94 (.98) \\
\strips 100 states & .943 (.954) & .820 (.801) & .98 (.98) & .964 (\textbf{1.0}) & .855 (\textbf{1.0}) & .69 (\textbf{1.0}) \\
\strips atoms & \textbf{1.0} (\textbf{1.0}) & .968 (.960) & .99 (.99) & .986 (.998) & .650 (.682) & .63 (.92) \\
\midrule
SB 50 states & \textbf{1.0} (\textbf{1.0}) & .999 (\textbf{1.0}) & \textbf{1.0} (\textbf{1.0}) & \textbf{1.0} (\textbf{1.0}) & .998 (.999) & \textbf{1.0} (\textbf{1.0}) \\
SB 100 states & \textbf{1.0} (\textbf{1.0}) & .999 (.999) & \textbf{1.0} (\textbf{1.0}) & \textbf{1.0} (\textbf{1.0}) & .998 (.999) & \textbf{1.0} (\textbf{1.0}) \\
SB atoms & \textbf{1.0} (\textbf{1.0}) & .994 (.996) & \textbf{1.0} (\textbf{1.0}) & \textbf{1.0} (\textbf{1.0}) & .997 (.997) & \textbf{1.0} (\textbf{1.0}) \\
\bottomrule
\end{tabular}

\vspace{0.05cm}

\begin{tabular}{@{}lrrrrrr@{}}
\toprule
 & \multicolumn{6}{c}{Maze} \\
 & \multicolumn{3}{c}{5x5} & \multicolumn{3}{c}{7x7} \\
\cmidrule(lr){2-4}\cmidrule(lr){5-7}
Model & Train & Test & Plan & Train & Test & Plan \\
\midrule
\strips 50 states & .906 (.944) & .514 (.580) & .63 (.70) & .897 (.932) & .659 (.721) & .44 (.50) \\
\strips 100 states & .862 (.868) & .361 (.375) & .34 (.37) & .784 (.798) & .428 (.442) & .15 (.19) \\
\strips atoms & .889 (.902) & .431 (.412) & .11 (.16) & .158 (.166) & .021 (.011) & .00 (.00) \\
\midrule
SB 50 states & \textbf{1.0} (\textbf{1.0}) & .997 (\textbf{1.0}) & .90 (.90) & \textbf{1.0} (\textbf{1.0}) & \textbf{1.0} (\textbf{1.0}) & .64 (.64) \\
SB 100 states & \textbf{1.0} (\textbf{1.0}) & .997 (.998) & \textbf{1.0} (\textbf{1.0}) & \textbf{1.0} (\textbf{1.0}) & .999 (\textbf{1.0}) & .99 (.99) \\
SB atoms & \textbf{1.0} (\textbf{1.0}) & \textbf{1.0} (.999) & .98 (.98) & \textbf{1.0} (\textbf{1.0}) & .958 (.967) & .98 (.98) \\
\bottomrule
\end{tabular}

\vspace{0.05cm}

\begin{tabular}{@{}lrrrrrr@{}}
\toprule
 & \multicolumn{6}{c}{Logistics} \\
 & \multicolumn{3}{c}{3 cities} & \multicolumn{3}{c}{5 cities} \\
\cmidrule(lr){2-4}\cmidrule(lr){5-7}
Model & Train & Test & Plan & Train & Test & Plan \\
\midrule
\strips 50 states & .985 (\textbf{1.0}) & .905 (\textbf{1.0}) & .95 (.97) & .988 (.999) & .821 (.891) & .54 (.73) \\
\strips 100 states & .988 (\textbf{1.0}) & .959 (\textbf{1.0}) & .91 (.97) & .989 (.999) & .851 (.895) & .38 (.37) \\
\strips atoms & .788 (.944) & .558 (.763) & .71 (.90) & .959 (\textbf{1.0}) & .902 (\textbf{1.0}) & .88 (\textbf{1.0}) \\
\midrule
SB 50 states & \textbf{1.0} (\textbf{1.0}) & .997 (.999) & \textbf{1.0} (\textbf{1.0})  & \textbf{1.0} (\textbf{1.0}) & .978 (.984) & \textbf{1.0} (\textbf{1.0}) \\
SB 100 states & \textbf{1.0} (\textbf{1.0}) & .997 (\textbf{1.0}) & \textbf{1.0} (\textbf{1.0})  & \textbf{1.0} (\textbf{1.0}) & .978 (.980) & \textbf{1.0} (\textbf{1.0}) \\
SB atoms & \textbf{1.0} (\textbf{1.0}) & .990 (.999) & \textbf{1.0} (\textbf{1.0})  & \textbf{1.0} (\textbf{1.0}) & .987 (.993) & \textbf{1.0} (\textbf{1.0}) \\
\bottomrule
\end{tabular}

\caption{\textbf{Main Results.} Comparison of \strips and SB transformers across five planning domains. We evaluate three configurations: \textit{init-state} encoding with 50 and 100 distinct initial states, and the \textit{init-atoms} encoding (\textit{atoms}). All configurations incorporate \textit{test-p} end setup actions. Metrics represent average training, test, and planning accuracy across three random seeds. Values in parentheses indicate results for the seed achieving the highest training accuracy.}
%\caption{\textbf{Main results.} We consider three configurations for the \strips and SB Transformers: \textit{init-state} encoding with 50 distinct initial states (\textit{50 states} row), \textit{init-state} with 100 initial states (\textit{100 states}), and \textit{init-atoms} encoding (\textit{atoms}). All configurations use end setup actions \textit{test-p}.
%For each model, we show its training (\textit{Train} column), test (\textit{Test}) and planning (\textit{Plan}) accuracy.
%Metrics correspond to averages across three random seeds. Between parenthesis, we report the value for the seed with the best training accuracy.
%}
\label{tab:main_results}
\end{table*}

\paragraph{Comparison with standard transformers.}
Table~\ref{tab:baselines} compares the SB Transformer against the two baseline models,
the standard transformer using softmax attention with sinusoidal positional encodings (Sinusoidal), and one  using Rotary Position Embeddings (RoPe).
While all models achieve high accuracy in  training (1st column) for classifying traces,
only the SB Transformer maintains high accuracy on the longer test traces (2nd column).
The last two columns focus on the accuracy of plans obtained from the  learned \strips models.

The results of the first two columns  indicate that while \emph{standard} transformers learn to classify the traces used in training,
they fail to generalize  to longer traces not seen during training. The  stick-breaking attention transformer, on the other hand,
bridges this  gap,  and yields robust generalization to new, long traces. 
%Furthermore, Table~\ref{tab:main_results} demonstrates that both the SB and \strips Transformers consistently
% achieve high test accuracy across all domains and problem sizes. 
Interestingly, the last two columns of the table show that while standard transformers show poor generalization over long test traces,
the symbolic \strips models that can be extracted from them, following the same method as the ones used by the SB Transformer,
are accurate and achieve nearly perfect generalization. More precisely, the \strips models obtained from standard transformers
are very  accurate when the training traces are short ($M_s$), but not when the training traces are long ($M_L$).
Once again, the SB Transformer yields perfectly accurate \strips models in both cases.

\Omit{
As shown in the Plan ($M_S$) column of Table~\ref{tab:baselines}, if a \strips model is extracted from the short training traces, the resulting symbolic model achieves perfect planning accuracy (near-perfect for \mz). This suggests that the baselines do learn the transition dynamics, although they cannot reliably predict them in longer sequences. 
In contrast, when the \strips\ model is extracted from longer test traces (Plan ($M_L$)), the performance of the baseline models collapses, while the SB Transformer maintains perfect accuracy (near-perfect for \mz).
}

%\paragraph{Transformer generalization can be improved using stick-breaking attention or by extracting a symbolic model.}
%Table~\ref{tab:baselines} shows that the \strips models $M_S$ extracted from the baseline transformers using the short, training traces result in perfect planning accuracy, whereas models $M_L$ extracted using the long, test traces result in 0\% accuracy.
%Therefore, there exist two alternatives for improving length-generalization in transformers: (1) removing positional embeddings and replacing softmax with stick-breaking attention (as the SB Transformer does), or (2) extracting a \strips model (on short traces) and using it in place of the transformer for both planning and predicting action applicability.

\paragraph{Comparison between the \strips\ Transformer and the SB Transformer.}
Table~\ref{tab:main_results} presents comprehensive results across all five planning domains (extended planning results are
provided in Section~\ref{appendix:extended_planning_results} of the Appendix). We observe that while both models achieve strong overall performance, the SB Transformer consistently outperforms the STRIPS Transformer. This occurs despite the latter incorporating additional symbolic built-in structure, such as a one-to-one mapping between parameters $\theta$ and action preconditions and effects.
Furthermore, the \strips\ Transformer exhibits greater variability across runs, and it occasionally fails to achieve high training accuracy despite
possessing sufficient expressivity to represent the domains. We hypothesize that this performance gap is primarily due to the  more
complex optimization process that has to be solved  via gradient descent. Nevertheless, Table~\ref{tab:main_results} indicates that when the model does attain high training accuracy, the corresponding test and planning accuracies tend to be high as well. For further results regarding the scaling behavior of the two architectures  with the amount of data available, see Section~\ref{appendix:experiments_different_sizes} of the Appendix.

%\paragraph{The \strips Transformer does \textit{not} improve over the SB Transformer.}
%Table~\ref{tab:main_results} shows that, although our two proposed models achieve good results overall, the SB Transformer outperforms the \strips Transformer, despite the latter incorporating additional  (e.g., a one-to-one mapping between parameters $\theta$ and action preconditions and effects).
%Moreover, the \strips Transformer exhibits greater variability across runs, sometimes attaining low training accuracy despite having enough expressive capacity to represent the ground-truth domain.
%Therefore, we believe the main reason behind this performance gap is that the strong, symbolic inductive biases encoded into the architecture of the \strips Transformer make optimization via gradient-descent difficult.
%Nonetheless, Table~\ref{tab:main_results} shows that, when it attains high training accuracy, then the corresponding test and planning accuracy also tend to be high.

\paragraph{Compositional Reasoning and Combinatorial Generalization.}
Listing~\ref{lst:pddl_domains} shows examples of \strips\ models extracted from our two proposed architectures after learning in \bwl\ under the \textit{init-atoms} configuration.
The model of the \strips Transformer is encoded in terms of atoms \texttt{pX} different to those of the ground-truth domain. To translate between learned atoms \texttt{pX} and ground-truth atoms $p \in F$, the \textit{init-p} and \textit{test-p} actions must be used. This means that only those \texttt{pX} appearing in initial states or goals can be aligned (i.e., translated) with the ground-truth domain.
The SB Transformer assumes a one-to-one correspondence between ground-truth atoms $p \in F$ and \textit{test-p} actions. Consequently, the learned model uses the same atoms as the ground-truth model. This comes at the cost of a larger set of \textit{test-p} actions and the expensive state probing process used to extract the model.
\iffalse
The \textit{init-p} and \textit{test-p} actions facilitate translation between learned atoms \texttt{pX} and ground-truth atoms $p \in F$ within the model $M=\tup{F,A}$. The SB Transformer assumes a one-to-one correspondence between each ground-truth atom $p \in F$ and a \textit{test-p} action; consequently, every learned atom \texttt{pX} is aligned, at the cost of the expensive state probing process. In contrast, the \strips\ Transformer employs a more sparse set of \textit{test-p} actions—specifically one per goal atom $p$—meaning only the subset of \texttt{pX} atoms appearing in initial states or goals can be directly aligned.
\fi

Despite these architectural differences, Table~\ref{tab:main_results} demonstrates that the \strips\ models extracted from both transformers successfully interface with off-the-shelf planners and symbolic heuristics (e.g., FF), yielding high overall planning accuracy. This robustness persists even for complex problems where solution plans require a high number of actions (see Section~\ref{appendix:extended_planning_results} in the Appendix). Furthermore, the use of \textit{test-p} actions allows these models to
generalize to an exponential number of potential goals,  distinct to those seen in training. Similarly, for the \textit{init-atoms} encoding, the models generalize to an exponential number of previously unseen initial states. We provide additional comparative results with different setup configurations for the initial and final states in
%initial and end setup configurations in
Section~\ref{appendix:symbolic_alignment} of the Appendix. % , showing that the  \textit{init-p} and \textit{test-p} setup actions improve the generalization abilities of the SB Transformer.

%In conclusion, both \strips\ and SB Transformers exhibit robust compositional reasoning by effectively planning with their learned symbolic models. The inclusion of \textit{init-p} and \textit{test-p} actions enables these architectures to generalize to an exponential number of unseen initial states and goals, respectively.

Overall, the experiments show that both the \strips Transformer and SB Transformer produce \strips models
that support effective  planning over the hidden \strips domain  and an exponential number of
new initial states and goals not seen during training.

%% file: conclusion.tex
We have shown that the \strips and SB transformers learn to classify positive and negative action traces in
\strips, generalizing to longer traces. Moreover,  propositional \strips  models can be extracted from them in order to support planning via off-the-shelf planners.
While the \strips Transformer follows an encoding of the \strips classification task in the B-RASP language (see Appendix), and thus has more built-in
structure, the SB Transformer is easier to train and  generalizes more robustly. Interestingly, standard transformers that do not use
stick-breaking attention do not generalize to long traces, although the symbolic \strips models that can be obtained from them do  when extracted on short traces.
In the  future, we plan to  extend our approach to learn lifted \strips models from action traces, as done by the symbolic  \sift algorithm
\cite{sift}. 
We note, however, that \sift is not suitable for learning propositional \strips models such as the ones
considered in this work, as it runs in time that is exponential in the number of action schemas, which in the propositional case
is equal to the number of actions (128 in \bwl, for example).

%% file: brasp.tex
\section{B-RASP Formulation}

\brasp is a Boolean version of the RASP language~\cite{rasp} which aims at providing an abstract representation of masked, hard-attention transformers, in order to facilitate their analysis~\cite{b-rasp}. A masked, hard-attention transformer with a particular set of parameter values can be compiled into an equivalent \brasp program, and vice versa.

In this section, we first review the basics of the \brasp language and then
present the \brasp program corresponding to the \strips transformer introduced
in Section~\ref{sec:strips_trans}.

A \brasp program uses Boolean vectors of length $n$ as its only datatype.
Throughout this section, positions in strings and vectors are indexed from
$0$ to $n-1$, we write $[n]=\{0,\ldots,n-1\}$. A \brasp program receives as
input a string $w=(w_0,\ldots,w_{n-1})$ composed of symbols $w_i$ from a finite alphabet $\Sigma$. The input string is encoded as a set of initial vectors $\mathcal{I}_\sigma$, one for each $\sigma\in\Sigma$, where $\mathcal{I}_\sigma(i)=1 \text{ iff } w_i=\sigma, i\in[n].$

If these $|\Sigma|$ initial vectors $\mathcal{I}_\sigma$ are denoted as $P_1, \ldots, P_{|\Sigma|}$, a \brasp program can be understood as defining new
vectors $P_t$ from previous vectors $P_1,\ldots,P_{t-1}$ through one of two
operations \footnote{Vector indices like $1,\ldots,t$ enumerate vectors
in the program, whereas position indices such as $i,j\in[n]$ refer to entries
of those vectors.}
\begin{description}
\item[Position-wise operations]: Each  element $P_t(i)$ of $P_t$ is defined as a  Boolean function  over elements from $\{P_1(i), \ldots, P_{t-1}(i)\}$, for $i\in[n]$.
\item[Attention operations]: Each $P_t(i)$ is defined as
\begin{equation}\label{eq:att-brasp}
P_t(i) := \blacktriangle_j [M(i,j),S(i,j)] \, V(i,j) : D(i),
\qquad i\in[n],
\end{equation}
where
\begin{itemize}
\item the operator $\blacktriangle_j$ is either $\blacktriangleleft_j$
(argmin) or $\blacktriangleright_j$ (argmax),
\item the \textit{mask predicate} $M(i,j)$ is $1$ (no masking), $(j<i)$
(strict future masking), or $(j>i)$ (strict past masking),
\item the \textit{score predicate} $S(i,j)$ and the \textit{value predicate}
$V(i,j)$ are each given by a possibly different Boolean formula over
$\{P_1(i),\ldots,P_{t-1}(i)\}\cup
 \{P_1(j),\ldots,P_{t-1}(j)\}$, and
\item the \textit{default value predicate} $D(i)$ is a Boolean formula over
$\{P_1(i),\ldots,P_{t-1}(i)\}$.
\end{itemize}
\end{description}

For each $i \in [n]$, let $j_i$ be the minimum index $j \in [n]$
if $\blacktriangle_j=\blacktriangleleft_j$, or the maximum index
$j \in [n]$ if $\blacktriangle_j=\blacktriangleright_j$, such that
$M(i,j)=1$ and $S(i,j)=1$, provided that such an index exists.
If such an index $j_i$ exists, then $P_t(i) := V(i,j_i)$.
Otherwise, $P_t(i) := D(i)$.
The intuition is that $P_t(i)$ is set to $V(i,j_i)$, a Boolean
formula over
$\{P_1(i), \dots, P_{t-1}(i)\} \cup
 \{P_1(j_i), \dots, P_{t-1}(j_i)\}$,
when position $i$ attends to position $j_i$. If no such position
$j_i$ exists, then $P_t(i)$ is set to $D(i)$, which is a Boolean
formula over $\{P_1(i), \dots, P_{t-1}(i)\}$ only.

% Vicenc: added this
Multi-head attention can be simulated by combining the outputs of multiple
attention operations of the form~\eqref{eq:att-brasp}. This is particularly
important for the compiled transformer, which can exploit this mechanism to
parallelize computations across attention heads at the same depth.

A \brasp program can be used either to determine whether an input string $w$
belongs to some language $L\subseteq\Sigma^*$, or to map input strings $w$
into output strings $w'\in\Gamma^*$ of the same length. In the first case,
$w$ is encoded as the set of input vectors above, and it is regarded as
accepted, or equivalently as belonging to $L$, if $Z(n-1)=1,$
where $Z$ is the last vector produced by the program. In the second case, $w$
is encoded as the set of input vectors above, and $w'$ is read from the last
$|\Gamma|$ vectors $Z_\gamma$ produced by the program, where $w'(i)=\gamma $ iff $Z_\gamma(i)=1$,$i\in[n]$, $\gamma\in\Gamma$.

\section{B-RASP Program for Propositional \strips}\label{sec:brasp_program}
We now describe the \brasp program equivalent to the \strips transformer
introduced in Section~\ref{sec:strips_trans}. This program computes
$f_M^{\brasp}$, which encodes the Boolean function $f_M$ that discriminates
positive from negative action traces $\tau$ of a given \strips model
$M=\tup{F,A}$.

%\hector{Notation: Let's use the following notations (feel free to improve fonts, etc):
%  $f_M$ is the trace classifier based on the model, $f_M^{\brasp}$ is the trace
%  classifier expressed in B-RASP, $f_T$ is the trace classifier learned from
% the set of traces $T$, and when we need to refer to the learnable parameters
%  $w$ of this function, write it as $f_T^w$ (or with $\theta$ instead of $w$ if
%  you prefer)}

In the \brasp program, the input trace is
$\tau=(a_0,\ldots,a_{n-1})$, with positions indexed by
$i,j\in[n]$. The program computes the
Boolean function $f_M$ defined in Section~5, namely
$f_M(\tau)=1$ iff $\tau$ is negative, or equivalently iff some action
$a_i$ is inapplicable given the prefix $(a_0,\ldots,a_{i-1})$, otherwise,
$f_M(\tau)=0$.

Input traces are encoded using $|A|$ initial vectors $I_a$ of size $n$,
one for each action $a\in A$, where
\[
I_a(i)=1 \quad \text{iff} \quad a_i=a,
\qquad i\in[n].
\]
For each atom $p\in F$ in $M$, we then define three vectors
$Q_p$, $K_p$, and $V_p$ that depend on the model $M$. These Boolean
vectors are the analogues of the structured query, key, and value
parameters of the \strips transformer.
%We define a \brasp program that computes $f_\text{IC}(\tau; M)$, where the input is an action trace $\tau = (a_1, \dots, a_n) \in A^n$.
%We start defining the following initial Boolean vectors for each possible action $a$ in the trace, $\mathcal{Q}_a(i) = 1$ iff $a_i=a$.
\begin{itemize}
  \item $Q_p(i)$ indicates whether $p\in\mathrm{pre}(a_i)$, and is
  analogous to Eqs.~\ref{eq:thq} and~\ref{eq:transQ}:
  \begin{align}\label{eq:rasp_query}
  Q_p(i) := \bigvee_{a \mid p \in \mathrm{pre}(a)} I_a(i).
  \end{align}

  \item $K_p(i)$ indicates whether $p$ appears in either
  $\mathrm{add}(a_i)$ or $\mathrm{del}(a_i)$, and is analogous to
  Eqs.~\ref{eq:thk} and~\ref{eq:transK}:
  \begin{align}\label{eq:rasp_key}
  K_p(i) := \bigvee_{a \mid p \in \mathrm{add}(a) \cup \mathrm{del}(a)}
  I_a(i).
  \end{align}

  \item $V_p(i)$ indicates whether $p\in\mathrm{del}(a_i)$, and is
  analogous to Eqs.~\ref{eq:thv} and~\ref{eq:transV}:
  \begin{align}\label{eq:rasp_value}
  V_p(i) := \bigvee_{a \mid p \in \mathrm{del}(a)} I_a(i).
  \end{align}
\end{itemize}
The remaining vectors are defined by Boolean operations over these three vectors, they do not depend explicitly on $M$:
\begin{itemize}
  \item $Y_p(i)$ indicates whether $p\in\mathrm{pre}(a_i)$ and the most
  recent action before position $i$ that affects $p$ deletes $p$. This is
  computed with an attention operation:
  \begin{align}\label{eq:rasp_attention}
  Y_p(i) :=
  \blacktriangleright_j~[j < i,\ Q_p(i) \land K_p(j)]\ V_p(j)\ :\ 0.
  \end{align}
\end{itemize}

Here, $Q_p(i)\land K_p(j)$ is the score predicate, $V_p(j)$ is the value
predicate, and $0$ is the default value predicate. Since
$\blacktriangleright_j$ selects the maximum index $j$ satisfying the mask
and score predicates, the operation attends to the most recent position
$j<i$ such that action $a_j$ affects $p$, but only when $p$ is a
precondition of $a_i$.

Thus, $Y_p(i)=1$ iff action $a_i$ has precondition $p$ and the most recent
preceding action that affects $p$ deletes it. In this case, $p$ is false
before applying $a_i$, and $a_i$ is inapplicable because of $p$. Conversely,
$Y_p(i)=0$ if $p$ is not a precondition of $a_i$, if no preceding action
affects $p$, or if the most recent preceding action that affects $p$ adds it.
This attention operation corresponds to Eqs.~\ref{eq:tr_scores}--\ref{eq:str_val}
in the \strips transformer.

%where we use $Q_p(i) \land K_p(j)$ as the score predicate, $V_p(j)$ as the value predicate, and $0$ as the default value predicate.
%Intuitively,  $Y_p(i)=1$ can be understood as a \emph{flag} of inapplicability, meaning that action $a_i$ contains a precondition $p$ that is made false by the last preceding  action
%in the trace $a_j$ that affects $p$ (remember that  $\blacktriangleright$ denotes max),
%and conversely,  $Y_p(i)=0$ means either that no action preceding $a_i$ in the trace affects~$p$ or $p$ is not in the precondition of $a_i$
%\emph{(default predicate)}, or that the last preceding action that affects~$p$, adds it.
%This attention operation corresponds to Eqs.~\ref{eq:tr_scores}--\ref{eq:str_val} in the \strips Transformer.

% We assume that all propositions $p \in P$ are true at the start of the trace. -> Hector: this doesn't seem to be needed

%Every action $a_i \in \tau$, $a_i$ will be a possible next action iff all of its preconditions $p \in \mathrm{pre}(a_i)$ are true. In that case, for each $p \in P$ we will have $F_p(i) = 0$. Therefore, we combine all the vectors $F_p(i)$ into a single vector $F(i)$ so that $F(i)=0$ iff $a_i$ is a possible next action:

Each action $a_i\in\tau$ is applicable iff all of its preconditions are true.
Equivalently, $a_i$ is inapplicable iff there exists some atom $p\in F$ such
that $Y_p(i)=1$. We therefore combine all the vectors $Y_p$ into a single
vector $Y$:
\begin{align}\label{eq:rasp_or_atoms}
Y(i) := \bigvee_{p\in F} Y_p(i).
\end{align}
Hence,
\[
Y(i)=1 \quad\text{iff action $a_i$ is inapplicable given the prefix }
(a_0,\ldots,a_{i-1}),
\]
and $Y(i)=0$ iff $a_i$ is applicable. This corresponds to
Eq.~\ref{eq:transf_output}. The vectors $Y_p$ for each $p\in F$ are computed
with separate attention operations, corresponding to separate attention heads,
and these computations are performed in parallel.

% , so the value of $Y(i)$ may depend on several actions $a_j, a_k, \ldots$ preceding $a_i$ in $\tau$. -> Hector> Not sure what thi sentence connects with the previous one
%\hector{*** This is important but we don't say anything of this the B-RASP review; namely, what's an ``an attention head'', having multiple attention heads, etc. Indeed, it's not even correct to say that the \strips transformer produces sequences of vectors; as some of these operations are in parallel/diff attention heads. We should clarify this well and be consistent (I'm not being fully consistent, and I'm not fully clear about this either. Need to do some thinking, but can't think while writing ;-). ***}

Finally, the trace $\tau$ is negative iff $Y(i)=1$ for some $i\in[n]$.
This is encoded by the output vector $Z$, which computes the Boolean OR of
$Y$:
\begin{align}\label{eq:rasp_output}
Z(i) :=
\blacktriangleright_j~[1,\ Y(j)]\ 1\ :\ 0.
\end{align}
Thus, $Z(i)=1$ iff at least one action in the trace $\tau$ is inapplicable.
The trace-level output of the \brasp program is given by the last position of this vector:
\begin{align}\label{eq:rasp-readout}
f_M^{\brasp}(\tau)=Z(n-1).
\end{align}

Equations~\ref{eq:rasp_output} and~\ref{eq:rasp-readout} correspond to
Eq.~\ref{eq:ftheta}. Therefore, this program computes the same function
$f_M(\tau)$ as in Section~5.
\begin{theorem}
Let $M=\tup{F,A}$ be a propositional \strips model, and let
$\tau=(a_0,\ldots,a_{n-1})$ be an action sequence over $A$. Then
$f_M^\brasp(\tau)=f_M(\tau)$.
\end{theorem}

Section~\ref{appendix:b_rasp_program} below illustrates the execution of this
\brasp program over two example traces $\tau$ to produce the Boolean values
$f_M^{\brasp}(\tau)$.

\Omit{
\noindent \textbf{Example.}
Fig.~\ref{fig:brasp-example}(a) shows a simple \strips model $M$ with three actions, $A=\{\aone, \atwo, \athree\}$, and three atoms, $F~=~\{\pone, \ptwo, \pthree\}$.
Positive traces alternate between $\aone$ and $\atwo$, with one or more occurrences of $\athree$ in-between.
Fig.~\ref{fig:brasp-example}(b) shows the execution of the \brasp program $f_M^{\brasp}$ on a positive trace $\tau^+$ (shown on top), while
Fig.~\ref{fig:brasp-example}(c) shows the execution for a  negative trace $\tau^-$ (shown on top as well).
It can be observed that $f_M^{\brasp}(\tau^+)=0$ and $f_M^{\brasp}(\tau^-)~=~1$, in agreement with $f_M$.
The computations are depicted in  five blocks separated by horizontal lines: the first block encodes the inputs, the inner three blocks
encode  computations that can be performed  in parallel from the input block, as  they correspond to three ``attention heads'',
and  the last block encodes the outputs from them. 
}

%% file: appendix.tex
    %% The folowing tables were generated using https://www.tablesgenerator.com/
%% The tables are in the following files, that can be loaded and edited:
%% accept_pone.tgn, accept_ptwo.tgn, accept_pthree.tgn
%% reject_pone.tgn, reject_ptwo.tgn, reject_pthree.tgn

\section{Example B-RASP Program}
\label{appendix:b_rasp_program}

Fig.~\ref{fig:brasp-example}(a) shows a simple \strips model $M$ with
three atoms, $F=\{\pone,\ptwo,\pthree\}$, three domain actions $A=\{\aone,\atwo,\athree\}$, and no setup actions. Positive traces alternate
between $\aone$ and $\atwo$, with one or more occurrences of $\athree$
in between.

Figs.~\ref{fig:brasp-example}(b) and~\ref{fig:brasp-example}(c) show the
execution of the \brasp program $f_M^{\brasp}$ on a positive trace
$\tau^+$ and a negative trace $\tau^-$, respectively. In both cases, the
trace is shown in the first row. The final output is read from the last
position of the vector $Z$, namely $Z(n-1)$ for a trace of length $n$.
Thus, the examples show that $f_M^{\brasp}(\tau^+)=0$ and
$f_M^{\brasp}(\tau^-)=1$, in agreement with the definition of $f_M$.

The computations are depicted in five blocks separated by horizontal
lines. The first block encodes the input vectors. The next three
blocks correspond to the computations associated with the atoms
$\pone$, $\ptwo$, and $\pthree$, respectively. These three blocks can be
computed in parallel, as they correspond to three attention heads. The
last block corresponds to the combined vector $Y$ and the output vector
$Z$.

\begin{figure*}[h]
\centering
\begin{subfigure}[t]{0.32\textwidth}
\centering
\raisebox{3cm}{
\begin{minipage}[t]{\linewidth}
\raggedright
\begin{center}
\textbf{Domain \domain}
\end{center}
\textbf{Atoms:}\\
\hspace{1cm} \pone, \ptwo, \pthree\\[0.5ex]
\textbf{Actions:}\\
\hspace{1cm} \aone:\\
\hspace{1.5cm} $\mathrm{pre}(\aone) = [\pone,\pthree]$\\
\hspace{1.5cm} $\mathrm{add}(\aone) = [\ptwo]$\\
\hspace{1.5cm} $\mathrm{del}(\aone) = [\pone,\pthree]$\\
\hspace{1cm} \atwo:\\
\hspace{1.5cm} $\mathrm{pre}(\atwo) = [\ptwo,\pthree]$\\
\hspace{1.5cm} $\mathrm{add}(\atwo) = [\pone]$\\
\hspace{1.5cm} $\mathrm{del}(\atwo) = [\ptwo,\pthree]$\\
\hspace{1cm} \athree:\\
\hspace{1.5cm} $\mathrm{pre}(\athree) = []$\\
\hspace{1.5cm} $\mathrm{add}(\athree) = [\pthree]$\\
\hspace{1.5cm} $\mathrm{del}(\athree) = []$\\
\end{minipage}
}
\caption{}
\end{subfigure}
\hfill
% Subfigure (b): Membership example
\begin{subfigure}[t]{0.32\textwidth}
\centering
\begin{tabular}{c|cccccc}
                   & $\aone$ & $\athree$ & $\athree$ & $\atwo$ & $\athree$ & $\aone$  \\
\midrule
$I_\aone$     & 1 & 0 & 0 & 0 & 0 & 1 \\
$I_\atwo$     & 0 & 0 & 0 & 1 & 0 & 0 \\
$I_\athree$     & 0 & 1 & 1 & 0 & 1 & 0 \\
\midrule
${Q}_{\pone}$          & 1 & 0 & 0 & 0 & 0 & 1 \\
${K}_{\pone}$          & 1 & 0 & 0 & 1 & 0 & 1 \\
${V}_{\pone}$          & 1 & 0 & 0 & 0 & 0 & 1 \\
${Y}_{\pone}$          & 0 & 0 & 0 & 0 & 0 & 0 \\
\midrule
${Q}_{\ptwo}$          & 0 & 0 & 0 & 1 & 0 & 0 \\
${K}_{\ptwo}$          & 1 & 0 & 0 & 1 & 0 & 1 \\
${V}_{\ptwo}$          & 0 & 0 & 0 & 1 & 0 & 0 \\
${Y}_{\ptwo}$          & 0 & 0 & 0 & 0 & 0 & 0 \\
\midrule
${Q}_{\pthree}$          & 1 & 0 & 0 & 1 & 0 & 1 \\
${K}_{\pthree}$          & 1 & 1 & 1 & 1 & 1 & 1 \\
${V}_{\pthree}$          & 1 & 0 & 0 & 1 & 0 & 1 \\
${Y}_{\pthree}$          & 0 & 0 & 0 & 0 & 0 & 0 \\
\midrule
${Y}$               & 0 & 0 & 0 & 0 & 0 & 0  \\
${Z}$               & 0 & 0 & 0 & 0 & 0 & \textbf{0}
\end{tabular}
\caption{}
\end{subfigure}
\hfill
% Subfigure (c): Non-membership example
\begin{subfigure}[t]{0.32\textwidth}
\centering
\begin{tabular}{c|cccccc}
                   & $\aone$ & $\athree$ & \textcolor{red}{$\aone$} & $\athree$ & $\atwo$ & \textcolor{red}{$\atwo$}  \\
\midrule
$I_\aone$     & 1 & 0 & 1 & 0 & 0 & 0 \\
$I_\atwo$     & 0 & 0 & 0 & 0 & 1 & 1 \\
$I_\athree$     & 0 & 1 & 0 & 1 & 0 & 0 \\
\midrule
${Q}_{\pone}$          & 1 & 0 & 1 & 0 & 0 & 0 \\
${K}_{\pone}$          & 1 & 0 & 1 & 0 & 1 & 1 \\
${V}_{\pone}$          & 1 & 0 & 1 & 0 & 0 & 0 \\
${Y}_{\pone}$          & 0 & 0 & \textcolor{red}{1} & 0 & 0 & 0\\
\midrule
${Q}_{\ptwo}$          & 0 & 0 & 0 & 0 & 1 & 1 \\
${K}_{\ptwo}$          & 1 & 0 & 1 & 0 & 1 & 1 \\
${V}_{\ptwo}$          & 0 & 0 & 0 & 0 & 1 & 1 \\
${Y}_{\ptwo}$          & 0 & 0 & 0 & 0 & 0 & \textcolor{red}{1} \\
\midrule
${Q}_{\pthree}$          & 1 & 0 & 1 & 0 & 1 & 1 \\
${K}_{\pthree}$          & 1 & 1 & 1 & 1 & 1 & 1 \\
${V}_{\pthree}$          & 1 & 0 & 1 & 0 & 1 & 1 \\
${Y}_{\pthree}$          & 0 & 0 & 0 & 0 & 0 & \textcolor{red}{1} \\
\midrule
${Y}$               & 0 & 0 & \textcolor{red}{1} & 0 & 0 & \textcolor{red}{1} \\
${Z}$               & 0 & 0 & 1 & 1 & 1 & \textcolor{red}{\textbf{1}} 
\end{tabular}
\caption{}
\end{subfigure}
\hfill
\caption{
Vectors produced by the \brasp program $f_M^\brasp$ for the two traces
$\tau^+$ and $\tau^-$ shown on top, drawn from the model $M$ shown on
the left, named $\domain$. The value of $f_M^\brasp(\tau)$ is given by
the last entry of the last vector $Z(5)$, in bold, for these traces of length
$n=6$.
For both traces, $f_M^\brasp(\tau)=f_M(\tau)$.
(a) The hidden \strips domain $\domain$.
(b) Vectors produced for computing $f_M^\brasp(\tau^+)$.
(c) Vectors produced for computing $f_M^\brasp(\tau^-)$.
The trace $\tau^-$ contains two inapplicable actions, marked in red:
the second occurrence of $\aone$, at position $2$, because the
precondition $\pone$ is false, giving $Y_{\pone}(2)=1$; and the last
occurrence of $\atwo$, at position $5$, because the preconditions
$\ptwo$ and $\pthree$ are false, giving
$Y_{\ptwo}(5)=Y_{\pthree}(5)=1$. Thus, $Y(2)=1$ and $Y(5)=1$,
meaning that $a_2=\aone$ and $a_5=\atwo$ are inapplicable actions.
Consequently, $\tau^-$ is a negative trace and the final output is
$Z(5)=1$.
}
\label{fig:brasp-example}
\end{figure*}

\FloatBarrier

\section{Illustration of \strips Transformer Computations}
\label{appendix:attention_computations}
%Figure~\ref{fig:brasp-example}, using the optimal parameterization $\theta=\theta_M$.
In this part of the Appendix, we illustrate the operations of the \strips Transformer (see Eqs. \ref{eq:transQ}-\ref{eq:ftheta}), using the same traces as Figure \ref{fig:brasp-example}, belonging to the \domain domain.
We use the optimal parameterization $\theta^*$ obtained from the ground-truth domain $M=\domain$, and set $Q_{p_l}$, $K_{p_l}, V_{p_l}$ using  Eqs.~\ref{eq:thq}-\ref{eq:thv}.
Figure \ref{fig:tstrips-example} shows the attention scores computed by each head, the application of strict future masking with stick-breaking attention, and the resulting output vectors $y_{p_l}$.
It can be observed that the transformer correctly predicts that the first trace (Fig. \ref{fig:tstrips-example}(a)) is positive and the second trace (Fig. \ref{fig:tstrips-example}(b)) is negative.

\begin{figure*}[h!]
\centering
\begin{subfigure}[t]{0.48\textwidth}
\centering
%\raisebox{3cm}{
\begin{minipage}[t]{\linewidth}
\scriptsize
\begin{tabular}{lllllllllll}
                      &                                & $S_{\pone}$                   &                                    &                                    &                                    &            &         &                       & $V_{\pone}$            & $y_{\pone}$              \\
                      &                                &                                    &                                    &                                    &                                    &            &         &                       &                          &                          \\
                      & \multicolumn{1}{l|}{$\aone$}   & $1\cdot 1$                         & $1\cdot 0$                         & $1\cdot 0$                         & \cellcolor[HTML]{C0C0C0}$1\cdot 1$ & $1\cdot 0$ &         & \multicolumn{1}{l|}{} & \multicolumn{1}{l|}{$0$} & \multicolumn{1}{l|}{$0$} \\
                      & \multicolumn{1}{l|}{$\athree$} & $0\cdot 1$                         & $0\cdot 0$                         & $0\cdot 0$                         & \cellcolor[HTML]{C0C0C0}$0\cdot 1$ &            &         & \multicolumn{1}{l|}{} & \multicolumn{1}{l|}{$0$} & \multicolumn{1}{l|}{$0$} \\
                      & \multicolumn{1}{l|}{$\atwo$}   & $0\cdot 1$                         & $0\cdot 0$                         & \cellcolor[HTML]{C0C0C0}$0\cdot 0$ &                                    &            &         & \multicolumn{1}{l|}{} & \multicolumn{1}{l|}{$0$} & \multicolumn{1}{l|}{$0$} \\
                      & \multicolumn{1}{l|}{$\athree$} & $0\cdot 1$                         & \cellcolor[HTML]{C0C0C0}$0\cdot 0$ &                                    &                                    &            &         & \multicolumn{1}{l|}{} & \multicolumn{1}{l|}{$0$} & \multicolumn{1}{l|}{$0$} \\
                      & \multicolumn{1}{l|}{$\athree$} & \cellcolor[HTML]{C0C0C0}$0\cdot 1$ &                                    &                                    &                                    &            &         & \multicolumn{1}{l|}{} & \multicolumn{1}{l|}{$1$} & \multicolumn{1}{l|}{$0$} \\
\multirow{-6}{*}{$i$} & \multicolumn{1}{l|}{$\aone$}   &                                    &                                    &                                    &                                    &            &         & \multicolumn{1}{l|}{} & \multicolumn{1}{l|}{$0$} & \multicolumn{1}{l|}{$0$} \\ \cline{3-8}
                      &                                & $\aone$                            & $\athree$                          & $\athree$                          & $\atwo$                            & $\athree$  & $\aone$ &                       &                          &                          \\
                      &                                & \multicolumn{6}{c}{$j$}                                                                                                                                                  &                       &                          &                         
\end{tabular}
\begin{tabular}{lllllllllll}
                      &                                & $S_{\ptwo}$                   &                                    &            &                                    &                                    &         &                       & $V_{\ptwo}$            & $y_{\ptwo}$              \\
                      &                                &                                    &                                    &            &                                    &                                    &         &                       &                          &                          \\
                      & \multicolumn{1}{l|}{$\aone$}   & $0\cdot 1$                         & $0\cdot 0$                         & $0\cdot 0$ & $0\cdot 1$                         & \cellcolor[HTML]{C0C0C0}$0\cdot 0$ &         & \multicolumn{1}{l|}{} & \multicolumn{1}{l|}{$0$} & \multicolumn{1}{l|}{$0$} \\
                      & \multicolumn{1}{l|}{$\athree$} & $0\cdot 1$                         & $0\cdot 0$                         & $0\cdot 0$ & \cellcolor[HTML]{C0C0C0}$0\cdot 1$ &                                    &         & \multicolumn{1}{l|}{} & \multicolumn{1}{l|}{$1$} & \multicolumn{1}{l|}{$0$} \\
                      & \multicolumn{1}{l|}{$\atwo$}   & \cellcolor[HTML]{C0C0C0}$1\cdot 1$ & $1\cdot 0$                         & $1\cdot 0$ &                                    &                                    &         & \multicolumn{1}{l|}{} & \multicolumn{1}{l|}{$0$} & \multicolumn{1}{l|}{$0$} \\
                      & \multicolumn{1}{l|}{$\athree$} & $0\cdot 1$                         & \cellcolor[HTML]{C0C0C0}$0\cdot 0$ &            &                                    &                                    &         & \multicolumn{1}{l|}{} & \multicolumn{1}{l|}{$0$} & \multicolumn{1}{l|}{$0$} \\
                      & \multicolumn{1}{l|}{$\athree$} & \cellcolor[HTML]{C0C0C0}$0\cdot 1$ &                                    &            &                                    &                                    &         & \multicolumn{1}{l|}{} & \multicolumn{1}{l|}{$0$} & \multicolumn{1}{l|}{$0$} \\
\multirow{-6}{*}{$i$} & \multicolumn{1}{l|}{$\aone$}   &                                    &                                    &            &                                    &                                    &         & \multicolumn{1}{l|}{} & \multicolumn{1}{l|}{$0$} & \multicolumn{1}{l|}{$0$} \\ \cline{3-8}
                      &                                & $\aone$                            & $\athree$                          & $\athree$  & $\atwo$                            & $\athree$                          & $\aone$ &                       &                          &                          \\
                      &                                & \multicolumn{6}{c}{$j$}                                                                                                                                                  &                       &                          &                         
\end{tabular}
\begin{tabular}{lllllllllll}
                      &                                & $S_{\pthree}$                 &                                    &                                    &                                    &                                    &         &                       & $V_{\pthree}$          & $y_{\pthree}$            \\
                      &                                &                                    &                                    &                                    &                                    &                                    &         &                       &                          &                          \\
                      & \multicolumn{1}{l|}{$\aone$}   & $1\cdot 1$                         & $1\cdot 1$                         & $1\cdot 1$                         & $1\cdot 1$                         & \cellcolor[HTML]{C0C0C0}$1\cdot 1$ &         & \multicolumn{1}{l|}{} & \multicolumn{1}{l|}{$0$} & \multicolumn{1}{l|}{$0$} \\
                      & \multicolumn{1}{l|}{$\athree$} & $0\cdot 1$                         & $0\cdot 1$                         & $0\cdot 1$                         & \cellcolor[HTML]{C0C0C0}$0\cdot 1$ &                                    &         & \multicolumn{1}{l|}{} & \multicolumn{1}{l|}{$1$} & \multicolumn{1}{l|}{$0$} \\
                      & \multicolumn{1}{l|}{$\atwo$}   & $1\cdot 1$                         & $1\cdot 1$                         & \cellcolor[HTML]{C0C0C0}$1\cdot 1$ &                                    &                                    &         & \multicolumn{1}{l|}{} & \multicolumn{1}{l|}{$0$} & \multicolumn{1}{l|}{$0$} \\
                      & \multicolumn{1}{l|}{$\athree$} & $0\cdot 1$                         & \cellcolor[HTML]{C0C0C0}$0\cdot 1$ &                                    &                                    &                                    &         & \multicolumn{1}{l|}{} & \multicolumn{1}{l|}{$0$} & \multicolumn{1}{l|}{$0$} \\
                      & \multicolumn{1}{l|}{$\athree$} & \cellcolor[HTML]{C0C0C0}$0\cdot 1$ &                                    &                                    &                                    &                                    &         & \multicolumn{1}{l|}{} & \multicolumn{1}{l|}{$1$} & \multicolumn{1}{l|}{$0$} \\
\multirow{-6}{*}{$i$} & \multicolumn{1}{l|}{$\aone$}   &                                    &                                    &                                    &                                    &                                    &         & \multicolumn{1}{l|}{} & \multicolumn{1}{l|}{$1$} & \multicolumn{1}{l|}{$0$} \\ \cline{3-8}
                      &                                & $\aone$                            & $\athree$                          & $\athree$                          & $\atwo$                            & $\athree$                          & $\aone$ &                       &                          &                          \\
                      &                                & \multicolumn{6}{c}{$j$}                                                                                                                                                                          &                       &                          &                         
\end{tabular}
\end{minipage}
%}
\caption{}
\end{subfigure}
\hfill
\begin{subfigure}[t]{0.48\textwidth}
\centering
\begin{minipage}[t]{\linewidth}
\scriptsize
\begin{tabular}{lllllllllll}
                      &                                & $S_{\pone}$                   &            &                                    &                                    &                                    &         &                       & $V_{\pone}$            & $y_{\pone}$                                      \\
                      &                                &                                    &            &                                    &                                    &                                    &         &                       &                          &                                                  \\
                      & \multicolumn{1}{l|}{$\atwo$}   & $0\cdot 1$                         & $0\cdot 0$ & $0\cdot 1$                         & $0\cdot 0$                         & \cellcolor[HTML]{C0C0C0}$0\cdot 1$ &         & \multicolumn{1}{l|}{} & \multicolumn{1}{l|}{$0$} & \multicolumn{1}{l|}{$0$}                         \\
                      & \multicolumn{1}{l|}{$\atwo$}   & $0\cdot 1$                         & $0\cdot 0$ & $0\cdot 1$                         & \cellcolor[HTML]{C0C0C0}$0\cdot 0$ &                                    &         & \multicolumn{1}{l|}{} & \multicolumn{1}{l|}{$0$} & \multicolumn{1}{l|}{$0$}                         \\
                      & \multicolumn{1}{l|}{$\athree$} & $0\cdot 1$                         & $0\cdot 0$ & \cellcolor[HTML]{C0C0C0}$0\cdot 1$ &                                    &                                    &         & \multicolumn{1}{l|}{} & \multicolumn{1}{l|}{$1$} & \multicolumn{1}{l|}{$0$}                         \\
                      & \multicolumn{1}{l|}{$\aone$}   & \cellcolor[HTML]{C0C0C0}$1\cdot 1$ & $1\cdot 0$ &                                    &                                    &                                    &         & \multicolumn{1}{l|}{} & \multicolumn{1}{l|}{$1$} & \multicolumn{1}{l|}{\cellcolor[HTML]{FD6864}$1$} \\
                      & \multicolumn{1}{l|}{$\athree$} & \cellcolor[HTML]{C0C0C0}$0\cdot 1$ &            &                                    &                                    &                                    &         & \multicolumn{1}{l|}{} & \multicolumn{1}{l|}{$1$} & \multicolumn{1}{l|}{$0$}                         \\
\multirow{-6}{*}{$i$} & \multicolumn{1}{l|}{$\aone$}   &                                    &            &                                    &                                    &                                    &         & \multicolumn{1}{l|}{} & \multicolumn{1}{l|}{$1$} & \multicolumn{1}{l|}{$0$}                         \\ \cline{3-8}
                      &                                & $\aone$                            & $\athree$  & $\aone$                            & $\athree$                          & $\atwo$                            & $\atwo$ &                       &                          &                                                  \\
                      &                                & \multicolumn{6}{c}{$j$}                                                                                                                                                  &                       &                          &                                                 
\end{tabular}
\begin{tabular}{lllllllllll}
                      &                                & $S_{\ptwo}$                   &                                    &                                    &            &                                    &         &                       & $V_{\ptwo}$            & $y_{\ptwo}$                                      \\
                      &                                &                                    &                                    &                                    &            &                                    &         &                       &                          &                                                  \\
                      & \multicolumn{1}{l|}{$\atwo$}   & $1\cdot 1$                         & $1\cdot 0$                         & $1\cdot 1$                         & $1\cdot 0$ & \cellcolor[HTML]{C0C0C0}$1\cdot 1$ &         & \multicolumn{1}{l|}{} & \multicolumn{1}{l|}{$1$} & \multicolumn{1}{l|}{\cellcolor[HTML]{FD6864}$1$} \\
                      & \multicolumn{1}{l|}{$\atwo$}   & $1\cdot 1$                         & $1\cdot 0$                         & \cellcolor[HTML]{C0C0C0}$1\cdot 1$ & $1\cdot 0$ &                                    &         & \multicolumn{1}{l|}{} & \multicolumn{1}{l|}{$0$} & \multicolumn{1}{l|}{$0$}                         \\
                      & \multicolumn{1}{l|}{$\athree$} & $0\cdot 1$                         & $0\cdot 0$                         & \cellcolor[HTML]{C0C0C0}$0\cdot 1$ &            &                                    &         & \multicolumn{1}{l|}{} & \multicolumn{1}{l|}{$0$} & \multicolumn{1}{l|}{$0$}                         \\
                      & \multicolumn{1}{l|}{$\aone$}   & $0\cdot 1$                         & \cellcolor[HTML]{C0C0C0}$0\cdot 0$ &                                    &            &                                    &         & \multicolumn{1}{l|}{} & \multicolumn{1}{l|}{$0$} & \multicolumn{1}{l|}{$0$}                         \\
                      & \multicolumn{1}{l|}{$\athree$} & \cellcolor[HTML]{C0C0C0}$0\cdot 1$ &                                    &                                    &            &                                    &         & \multicolumn{1}{l|}{} & \multicolumn{1}{l|}{$0$} & \multicolumn{1}{l|}{$0$}                         \\
\multirow{-6}{*}{$i$} & \multicolumn{1}{l|}{$\aone$}   &                                    &                                    &                                    &            &                                    &         & \multicolumn{1}{l|}{} & \multicolumn{1}{l|}{$0$} & \multicolumn{1}{l|}{$0$}                         \\ \cline{3-8}
                      &                                & $\aone$                            & $\athree$                          & $\aone$                            & $\athree$  & $\atwo$                            & $\atwo$ &                       &                          &                                                  \\
                      &                                & \multicolumn{6}{c}{$j$}                                                                                                                                                  &                       &                          &                                                 
\end{tabular}
\begin{tabular}{lllllllllll}
                      &                                & $S_{\pthree}$                 &                                    &                                    &                                    &                                    &         &                       & $V_{\pthree}$          & $y_{\pthree}$                                    \\
                      &                                &                                    &                                    &                                    &                                    &                                    &         &                       &                          &                                                  \\
                      & \multicolumn{1}{l|}{$\atwo$}   & $1\cdot 1$                         & $1\cdot 1$                         & $1\cdot 1$                         & $1\cdot 1$                         & \cellcolor[HTML]{C0C0C0}$1\cdot 1$ &         & \multicolumn{1}{l|}{} & \multicolumn{1}{l|}{$1$} & \multicolumn{1}{l|}{\cellcolor[HTML]{FD6864}$1$} \\
                      & \multicolumn{1}{l|}{$\atwo$}   & $1\cdot 1$                         & $1\cdot 1$                         & $1\cdot 1$                         & \cellcolor[HTML]{C0C0C0}$1\cdot 1$ &                                    &         & \multicolumn{1}{l|}{} & \multicolumn{1}{l|}{$0$} & \multicolumn{1}{l|}{$0$}                         \\
                      & \multicolumn{1}{l|}{$\athree$} & $0\cdot 1$                         & $0\cdot 1$                         & \cellcolor[HTML]{C0C0C0}$0\cdot 1$ &                                    &                                    &         & \multicolumn{1}{l|}{} & \multicolumn{1}{l|}{$1$} & \multicolumn{1}{l|}{$0$}                         \\
                      & \multicolumn{1}{l|}{$\aone$}   & $1\cdot 1$                         & \cellcolor[HTML]{C0C0C0}$1\cdot 1$ &                                    &                                    &                                    &         & \multicolumn{1}{l|}{} & \multicolumn{1}{l|}{$0$} & \multicolumn{1}{l|}{$0$}                         \\
                      & \multicolumn{1}{l|}{$\athree$} & \cellcolor[HTML]{C0C0C0}$0\cdot 1$ &                                    &                                    &                                    &                                    &         & \multicolumn{1}{l|}{} & \multicolumn{1}{l|}{$1$} & \multicolumn{1}{l|}{$0$}                         \\
\multirow{-6}{*}{$i$} & \multicolumn{1}{l|}{$\aone$}   &                                    &                                    &                                    &                                    &                                    &         & \multicolumn{1}{l|}{} & \multicolumn{1}{l|}{$0$} & \multicolumn{1}{l|}{$0$}                         \\ \cline{3-8}
                      &                                & $\aone$                            & $\athree$                          & $\aone$                            & $\athree$                          & $\atwo$                            & $\atwo$ &                       &                          &                                                  \\
                      &                                & \multicolumn{6}{c}{$j$}                                                                                                                                                                          &                       &                          &                                                 
\end{tabular}
\end{minipage}
\caption{}
\end{subfigure}
\hfill
\caption{
Self-attention computations for
(a) the valid trace $(\aone,\athree,\athree,\atwo,\athree,\aone)$ and
(b) the invalid trace $(\aone,\athree,\aone,\athree,\atwo,\atwo)$ in the
\strips domain $\domain$, see Fig.~\ref{fig:brasp-example}, using the
optimal parameters $\theta_M$.
Each attention head $\mathrm{att}_{p_l}$, where
$p_l\in\{\pone,\ptwo,\pthree\}$, shows the score matrix
$S_{p_l}=Q_{p_l}\cdot K_{p_l}^\top$, the strict future mask
(blank cells are masked), and the resulting output vector
$y_{p_l}=S'_{p_l}\cdot V_{p_l}$ after stick-breaking attention.
Since every score $S_{p_l}(i,j)$ is either $0$ or $1$, stick-breaking
attention behaves like hard attention: whenever a row contains at least
one unmasked score equal to $1$, the rightmost such score receives
normalized weight $1$, and all other positions receive weight $0$.
Rows with no unmasked score equal to $1$ produce the zero output for that
head. Gray shading illustrates the rightmost-predecessor computation;
only highlighted cells whose displayed score is $1$ receive normalized
weight $1$ after stick-breaking attention.
The output $y_{p_l}(i)$ is an atom-wise inapplicability flag:
$y_{p_l}(i)=1$ iff action $a_i$ is inapplicable because precondition
$p_l$ is false; otherwise $y_{p_l}(i)=0$.
For the invalid trace in (b), $y_{\pone}(2)=1$, meaning that
$a_2=\aone$ is inapplicable because $\pone$ is false. Moreover,
$y_{\ptwo}(5)=y_{\pthree}(5)=1$, meaning that $a_5=\atwo$ is
inapplicable because both $\ptwo$ and $\pthree$ are false. Therefore,
the trace is negative.
}
\label{fig:tstrips-example}
\end{figure*}

\FloatBarrier

\section{Experiments with Different Dataset Sizes}
\label{appendix:experiments_different_sizes}

\begin{figure*}[h]
    \centering
    \begin{subfigure}[t]{0.33\textwidth}
        \centering
        \includegraphics[width=\textwidth]{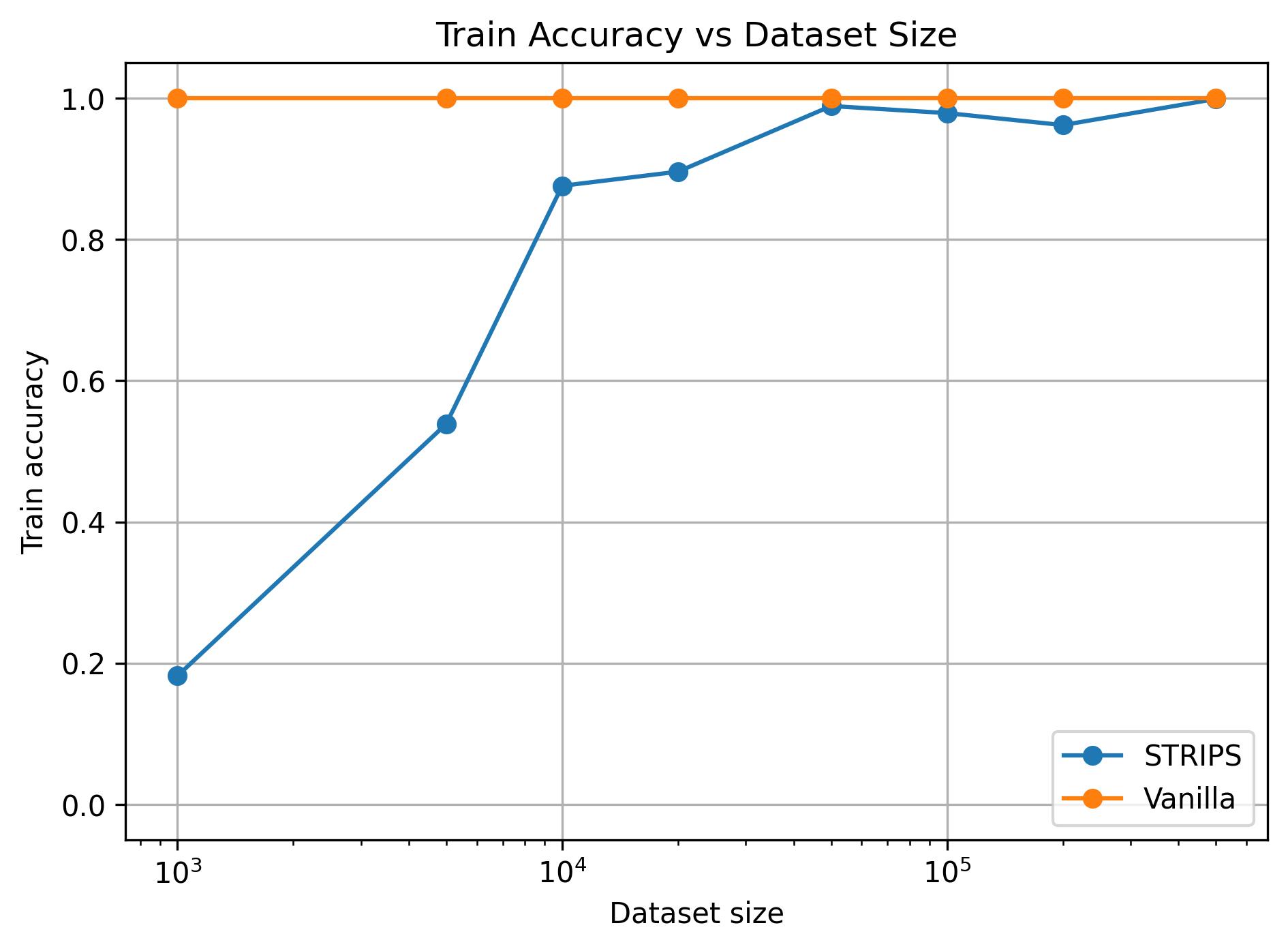}
        \caption{Train accuracy}
        \label{fig:train_acc_plot.jpg}
    \end{subfigure}
    \hfill
    \begin{subfigure}[t]{0.33\textwidth}
        \centering
        \includegraphics[width=\textwidth]{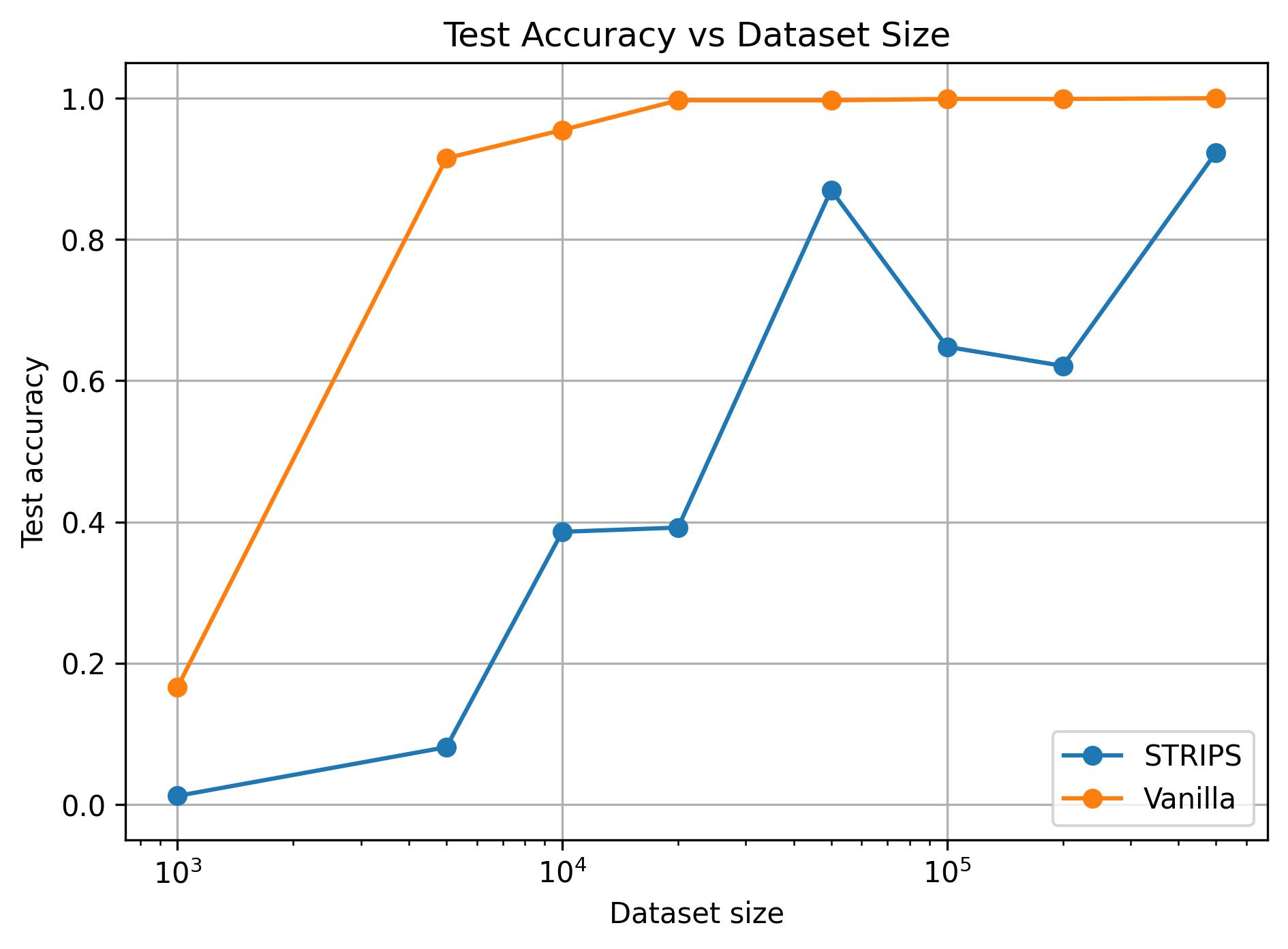}
        \caption{Test accuracy}
        \label{fig:test_acc_plot.jpg}
    \end{subfigure}
    \hfill
    \begin{subfigure}[t]{0.33\textwidth}
        \centering
        \includegraphics[width=\textwidth]{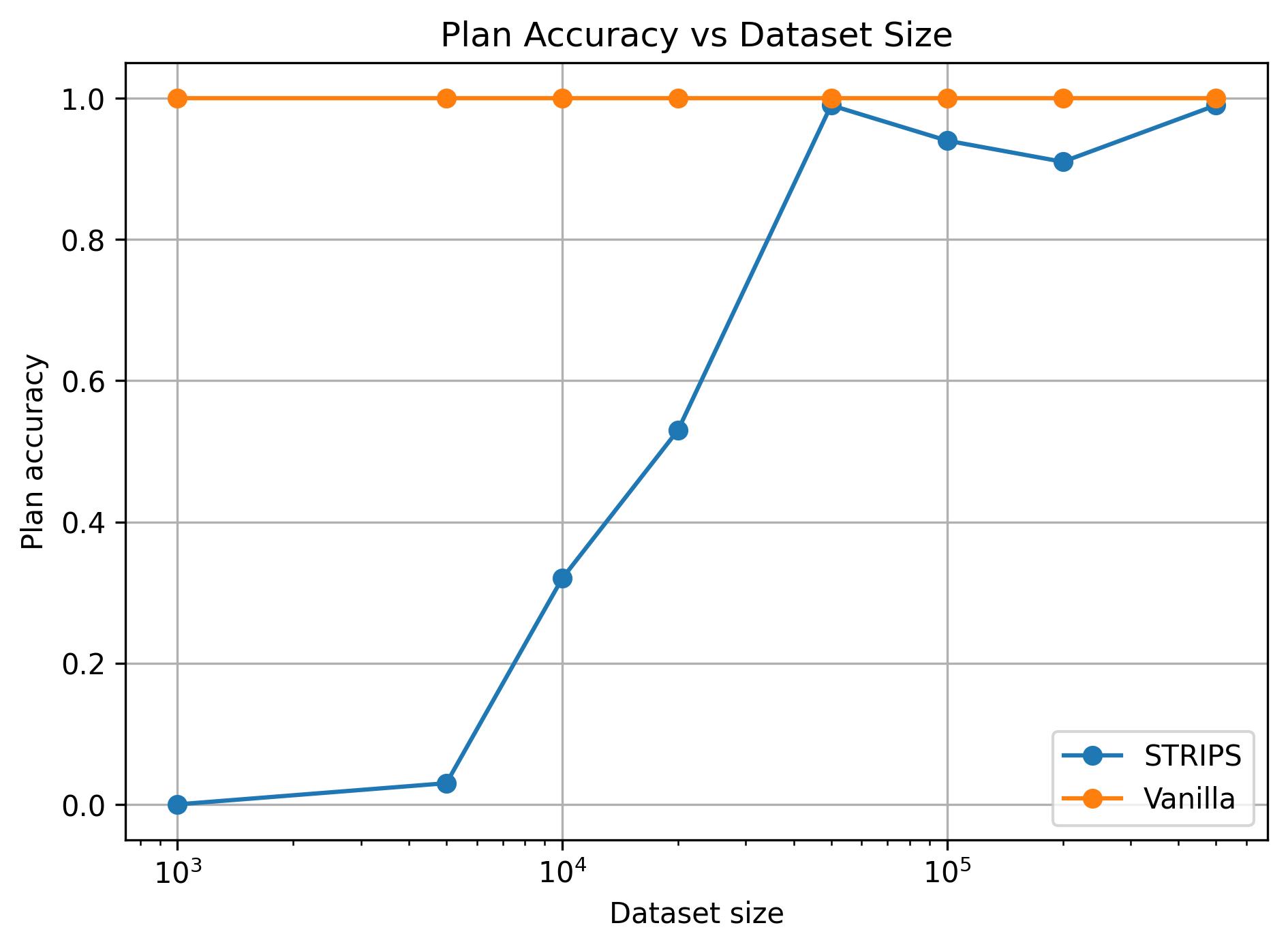}
        \caption{Planning accuracy}
        \label{fig:plan_acc_plot.jpg}
    \end{subfigure}

    \caption{
    \textbf{Experiments with different dataset sizes.} We measure the training, test and planning accuracy for the \strips and SB transformers when trained on datasets of different sizes (i.e., with different number of traces). Experiments are run in \bwl. For each model and dataset size, we use three random seeds and plot the metrics of the seed with the best training accuracy.
    }
    \label{fig:dataset-size-accuracy}
\end{figure*}

Fig.~\ref{fig:dataset-size-accuracy} shows the training, test, and planning
accuracy of the SB and \strips transformers when trained on datasets of
different sizes in \bwl.

We observe that the SB Transformer performs well even with small datasets:
it needs only $10^3$ training traces to achieve perfect planning accuracy,
and $5\cdot 10^3$ traces to exceed $90\%$ test accuracy. However, it needs
$10^5$ training traces to reach a test accuracy of $99.9\%$, so we use this
dataset size in the rest of our experiments.

In contrast, the \strips Transformer performs poorly on small datasets,
especially with fewer than $5\cdot 10^4$ traces. Interestingly, training
accuracy is also lower on smaller datasets. We hypothesize that training on
a larger and more diverse dataset helps the optimization process by reducing
the likelihood of getting stuck in local optima, yielding models with better
training, test, and planning accuracy.

\FloatBarrier

\section{Symbolic Alignment Experiments}
\label{appendix:symbolic_alignment}

Table~\ref{tab:symbolic_alignment} compares the performance of the SB and
\strips transformers under different setup configurations in \bwl. We vary
the initial setup actions, using either \textit{init-atoms} or
\textit{init-state} with one or 100 initial states, and the end setup
configuration, with or without appending \textit{test-p} actions to the
traces.

For every initial setup configuration, adding \textit{test-p} actions improves
the test accuracy of the SB Transformer. This is notable because
\textit{test-p} actions make the prediction task harder: the transformer must
classify not only domain actions, but also end setup actions that query the
truth values of individual atoms. We hypothesize that these actions provide a
form of \textit{symbolic alignment}. To predict the applicability of each
\textit{test-p} action, the transformer must track the truth value of the
corresponding atom $p$ as the trace is executed. This encourages a more
compositional representation in which atoms of the ground-truth domain are
tracked separately, which in turn improves generalization to longer test
traces.

This effect is especially clear for the \textit{init-state} configuration with
100 states (second row). Without \textit{test-p} actions, the SB Transformer achieves high
training accuracy but substantially lower test accuracy, suggesting that it
overfits to the training traces when many initial states are provided without
information about their internal atom structure. Adding \textit{test-p}
actions raises test accuracy from $.871$ to $.985$ on average.

The results also suggest that \textit{init-p} actions contribute to symbolic
alignment. Regardless of whether \textit{test-p} actions are used, the SB
Transformer obtains higher test accuracy with \textit{init-atoms} than with
\textit{init-state} using 100 states, even though \textit{init-atoms} requires
generalization to unseen initial states. This indicates that exposing the
atomic structure of initial states through \textit{init-p} actions helps the
model generalize.

We do not observe the same pattern for the \strips Transformer. This is
expected, since the \strips Transformer is symbolically aligned by design:
there is a direct correspondence between its parameters $\theta$ and the
preconditions and effects of the encoded \strips model $M$. Consequently,
additional setup actions do not consistently improve performance, and in some
configurations they make optimization harder.

%We hypothesize that \textit{test-p} actions act as a form of \textit{symbolic alignment} for the model: in order to correctly predict the applicability of each \textit{test-p} action, the transformer is forced to track the truth value of the $p$ atom as the trace actions are executed. Therefore, the transformer is encouraged to learn a compositional, disentangled model that separately tracks each atom $p$ in the ground-truth domain. Such a model generalizes better to longer, test traces, as Table~\ref{tab:symbolic_alignment} shows.

%This symbolic alignment effect is especially prominent for the \textit{init-state} configuration with 100 states, which shows that training the transformer on a large number of initial states without providing information about their inner structure (i.e., using no \textit{init-p} or \textit{test-p} actions) results in overfitting.

%Lastly, regardless of the use of end setup actions, the test accuracy of the SB Transformer improves when using \textit{init-atoms} actions instead of \textit{init-state} actions (for 100 states), despite the former requiring generalization to unseen initial states. This means that both \textit{init-p} and \textit{test-p} actions serve to symbolically align the model.

%Finally, we do not observe the same symbolic alignment effects for the \strips Transformer. This makes sense, since the \strips Transformer is symbolically aligned \textit{by design}, i.e., there is a one-to-one correspondence between its parameters $\theta$ and the action preconditions and effects of the encoded \strips model $M$.

\begin{table*}[h]
\centering
\small

\begin{tabular}{@{}lrrrr@{}}
\toprule
& \multicolumn{2}{c}{Without \textit{test-p} actions} & \multicolumn{2}{c}{With \textit{test-p} actions} \\
\cmidrule(lr){2-3}\cmidrule(lr){4-5}
Model & Train & Test & Train & Test \\
\midrule
\strips 1 state    & .984 (.991) & .590 (.762) & .729 (.903) & .348 (.535) \\
\strips 100 states & .973 (.976) & .617 (.660) & \textbf{1.0} (\textbf{1.0}) & .996 (.996) \\
\strips atoms     & .435 (.801) & .099 (.225) & .544 (.979) & .286 (.648) \\
\midrule
SB 1 state    & \textbf{1.0} (\textbf{1.0}) & .961 (.966) & \textbf{1.0} (\textbf{1.0}) & .969 (.979) \\
SB 100 states & .998 (.999) & .871 (.882) & \textbf{1.0} (\textbf{1.0}) & .985 (.994) \\
SB atoms     & \textbf{1.0} (\textbf{1.0}) & .951 (.926) & \textbf{1.0} (\textbf{1.0}) & .998 (.999) \\
\bottomrule
\end{tabular}

\caption{\textbf{Symbolic alignment results.} 
Symbolic alignment experiments in \bwl. We compare initial setup
configurations, using either \textit{init-state} with one or 100 initial
states or \textit{init-atoms}, with and without end setup actions
\textit{test-p}. Values report average training and test accuracy across
three random seeds; values in parentheses correspond to the seed with the
highest training accuracy. Adding \textit{test-p} actions consistently
improves the test accuracy of the SB Transformer, while the effect is not
consistent for the \strips Transformer.
%We measure the training and test accuracy of the SB and \strips transformers in \bwl for different initial setup actions (\textit{init-atoms} and \textit{init-state} with one and 100 initial states) and for different end setup configurations (with and without predicting \textit{test-p} actions). Since end setup actions are required for planning, we cannot measure planning accuracy. 
%For the SB transformer, we observe that predicting the goal atoms acts as a slight ``symbolic alignment'' which helps prevent overfitting on the training dataset. This results in a slight increase in test accuracy from \textit{no goal} to \textit{goal atoms} across all init state configurations. This effect is most prominent for the \textit{100 states} row, showing that using $init-p$ actions instead (i.e., performing symbolic alignment also on the initial state) also helps prevent overfitting.
}
\label{tab:symbolic_alignment}
\end{table*}

\FloatBarrier

\section{Extended Planning Results}
\label{appendix:extended_planning_results}

Tables~\ref{tab:extended_planning_part1} and~\ref{tab:extended_planning_part2}
provide additional information about the planning experiments. Overall, the
\strips and SB transformers produce extracted \strips models that allow
off-the-shelf planners to solve problems efficiently, with low planning times
and relatively few generated states across domains and problem sizes. This
holds even when the resulting plans contain many actions, as shown by the
\textit{Mean len} and \textit{Max len} columns.

Regarding planning errors, the model extracted from the \strips Transformer sometimes produces inapplicable plans (\textit{Inapplicable}) or fails to produce a plan at all (\textit{Unsolvable}).  The former case is caused by false positives in action applicability prediction: the learned model predicts an action as applicable although it is
inapplicable in the ground-truth domain.
The latter case is often caused by false negatives: the learned model predicts an action as inapplicable although it is applicable in the ground-truth domain. If such an action is needed to
solve the problem, the planner may fail to find a plan. The \textit{Bad goal}
column is zero or close to zero in almost all domains, suggesting that the
\strips Transformer usually predicts the applicability of \textit{test-p}
actions correctly, and hence recognizes goals correctly.

%The latter case is often caused by \textit{false negatives} in action applicability prediction, i.e., the learned model predicting an action as inapplicable when it is actually applicable in the ground-truth domain. Then, if such an action is required to solve the corresponding problem, no plan will be found. Also, we note that the \textit{Bad goal} column contains zero or near-zero values for all domains, meaning that the \strips Transformer correctly predicts the applicability of \textit{test-p} actions (i.e., it correctly recognizes goals).
%For the SB Transformer, the few errors observed all correspond to \textit{false negatives} in action applicability, resulting in unsolved problems.

For the SB Transformer, the few failures observed correspond to unsolved
problems. This suggests that these failures are due to false negatives in
action applicability, which remove transitions that are needed to solve the
problem, rather than to false positives that would lead to inapplicable plans.

\begin{table*}[!h]
\centering
\small

\begin{tabular}{@{}lrrrrrrrr@{}}
\toprule
 & \multicolumn{8}{c}{Blocksworld 5 blocks} \\
\cmidrule(lr){2-9}
Model & Correct & Inapplicable & Bad goal & Unsolved & Mean len & Max len & Time (s) & \# states \\
\midrule
\strips 50 states & .93 (\textbf{1.0}) & .06 (.00) & .01 (.00) & .01 (.00) & 12 (12) & 25 (24) & .0024 (.0024) & 71 (69) \\
\strips 100 states & .98 (\textbf{1.0}) & .00 (.00) & .02 (.00) & .00 (.00) & 12 (12) & 26 (26) & .0024 (.0022) & 72 (68) \\
\strips atoms & .65 (\textbf{1.0}) & .02 (.00) & .00 (.00) & .34 (.00) & 12 (12) & 29 (26) & .0029 (.0018) & 136 (56) \\
\midrule
SB 50 states & \textbf{1.0} (\textbf{1.0}) & .00 (.00) & .00 (.00) & .00 (.00) & 12 (12) & 26 (26) & .0016 (.0014) & 43 (43) \\
SB 100 states & \textbf{1.0} (\textbf{1.0}) & .00 (.00) & .00 (.00) & .00 (.00) & 12 (12) & 28 (28) & .0014 (.0014) & 45 (45) \\
SB atoms & \textbf{1.0} (\textbf{1.0}) & .00 (.00) & .00 (.00) & .00 (.00) & 12 (12) & 26 (26) & .0016 (.0015) & 45 (45) \\
\bottomrule
\end{tabular}

\vspace{0.1cm}

\begin{tabular}{@{}lrrrrrrrr@{}}
\toprule
 & \multicolumn{8}{c}{Blocksworld 8 blocks} \\
\cmidrule(lr){2-9}
Model & Correct & Inapplicable & Bad goal & Unsolved & Mean len & Max len & Time (s) & \# states \\
\midrule
\strips 50 states & .94 (.98) & .03 (.02) & .00 (.00) & .02 (.00) & 27 (26) & 56 (52) & .0246 (.0203) & 802 (666) \\
\strips 100 states & .91 (.96) & .09 (.04) & .00 (.00) & .00 (.00) & 26 (27) & 55 (54) & .0300 (.0252) & 814 (845) \\
\strips atoms & .53 (.94) & .00 (.00) & .01 (.00) & .45 (.06) & 29 (28) & 63 (62) & .2813 (.7614) & 11457 (31050) \\
\midrule
SB 50 states & \textbf{1.0} (\textbf{1.0}) & .00 (.00) & .00 (.00) & .00 (.00) & 27 (27) & 62 (62) & .0089 (.0076) & 317 (317) \\
SB 100 states & \textbf{1.0} (\textbf{1.0}) & .00 (.00) & .00 (.00) & .00 (.00) & 26 (26) & 56 (56) & .0073 (.0073) & 295 (295) \\
SB atoms & \textbf{1.0} (\textbf{1.0}) & .00 (.00) & .00 (.00) & .00 (.00) & 27 (27) & 54 (54) & .0076 (.0075) & 318 (318) \\
\bottomrule
\end{tabular}

\vspace{0.1cm}

\begin{tabular}{@{}lrrrrrrrr@{}}
\toprule
 & \multicolumn{8}{c}{Ferry 5 cars} \\
\cmidrule(lr){2-9}
Model & Correct & Inapplicable & Bad goal & Unsolved & Mean len & Max len & Time (s) & \# states \\
\midrule
\strips 50 states & .95 (\textbf{1.0}) & .00 (.00) & .00 (.00) & .05 (.00) & 15 (16) & 21 (22) & .0023 (.0024) & 77 (76) \\
\strips 100 states & \textbf{1.0} (\textbf{1.0}) & .00 (.00) & .00 (.00) & .00 (.00) & 14 (15) & 21 (21) & .0022 (.0025) & 69 (84) \\
\strips atoms & .92 (\textbf{1.0}) & .00 (.00) & .00 (.00) & .08 (.00) & 14 (14) & 20 (20) & .0018 (.0016) & 55 (44) \\
\midrule
SB 50 states & \textbf{1.0} (\textbf{1.0}) & .00 (.00) & .00 (.00) & .00 (.00) & 14 (14) & 19 (19) & .0016 (.0015) & 45 (45) \\
SB 100 states & \textbf{1.0} (\textbf{1.0}) & .00 (.00) & .00 (.00) & .00 (.00) & 14 (14) & 19 (19) & .0015 (.0015) & 44 (44) \\
SB atoms & \textbf{1.0} (\textbf{1.0}) & .00 (.00) & .00 (.00) & .00 (.00) & 14 (14) & 20 (20) & .0016 (.0015) & 43 (43) \\
\bottomrule
\end{tabular}

\vspace{0.1cm}

\begin{tabular}{@{}lrrrrrrrr@{}}
\toprule
 & \multicolumn{8}{c}{Ferry 8 cars} \\
\cmidrule(lr){2-9}
Model & Correct & Inapplicable & Bad goal & Unsolved & Mean len & Max len & Time (s) & \# states \\
\midrule
\strips 50 states & .98 (.93) & .02 (.05) & .00 (.00) & .01 (.02) & 26 (27) & 34 (36) & .0277 (.0535) & 908 (1802) \\
\strips 100 states & .88 (.97) & .01 (.03) & .00 (.00) & .11 (.00) & 27 (27) & 39 (38) & .0215 (.0294) & 735 (1067) \\
\strips atoms & .28 (.70) & .20 (.00) & .00 (.00) & .52 (.30) & 16 (25) & 19 (31) & .0043 (.0064) & 114 (200) \\
\midrule
SB 50 states & \textbf{1.0} (\textbf{1.0}) & .00 (.00) & .00 (.00) & .00 (.00) & 24 (24) & 32 (32) & .0040 (.0040) & 113 (113) \\
SB 100 states & \textbf{1.0} (\textbf{1.0}) & .00 (.00) & .00 (.00) & .00 (.00) & 24 (24) & 31 (31) & .0044 (.0040) & 111 (111) \\
SB atoms & \textbf{1.0} (\textbf{1.0}) & .00 (.00) & .00 (.00) & .00 (.00) & 24 (24) & 31 (31) & .0041 (.0043) & 113 (113) \\
\bottomrule
\end{tabular}

\caption{\textbf{Extended planning results}. For each model and configuration, we provide the percentage of planning problems for which a valid plan is found (\textit{Correct} column), where the plan contains one or several inapplicable actions (\textit{Inapplicable}), where the plan is applicable but does not achieve the problem goal (\textit{Bad goal}), and the percentage of problems were no plan was found (\textit{Unsolved}), due to either exhausting the search space or exceeding $10^6$ generated states. Also, we measure the average and maximum plan length (\textit{Mean len} and \textit{Max len}), average planning time in seconds (\textit{Time (s)}), and average number of generated states (\textit{\# states}). For these last four columns, we only consider problems that were successfully solved.}
\label{tab:extended_planning_part1}
\end{table*}

\begin{table*}[t]
\centering
\small

\begin{tabular}{@{}lrrrrrrrr@{}}
\toprule
 & \multicolumn{8}{c}{Npuzzle 5 tiles} \\
\cmidrule(lr){2-9}
Model & Correct & Inapplicable & Bad goal & Unsolved & Mean len & Max len & Time (s) & \# states \\
\midrule
\strips 50 states & .98 (\textbf{1.0}) & .00 (.00) & .02 (.00) & .00 (.00) & 15 (15) & 44 (38) & .0028 (.0027) & 81 (83) \\
\strips 100 states & .98 (.98) & .00 (.01) & .02 (.01) & .00 (.00) & 17 (18) & 47 (58) & .0027 (.0030) & 90 (105) \\
\strips atoms & .99 (.99) & .00 (.00) & .00 (.00) & .01 (.01) & 17 (17) & 44 (42) & .0023 (.0021) & 85 (72) \\
\midrule
SB 50 states & \textbf{1.0} (\textbf{1.0}) & .00 (.00) & .00 (.00) & .00 (.00) & 16 (16) & 54 (54) & .0019 (.0019) & 59 (59) \\
SB 100 states & \textbf{1.0} (\textbf{1.0}) & .00 (.00) & .00 (.00) & .00 (.00) & 16 (16) & 44 (44) & .0019 (.0019) & 59 (59) \\
SB atoms & \textbf{1.0} (\textbf{1.0}) & .00 (.00) & .00 (.00) & .00 (.00) & 17 (17) & 44 (44) & .0020 (.0020) & 61 (61) \\
\bottomrule
\end{tabular}

\vspace{0.1cm}

\begin{tabular}{@{}lrrrrrrrr@{}}
\toprule
 & \multicolumn{8}{c}{Npuzzle 8 tiles} \\
\cmidrule(lr){2-9}
Model & Correct & Inapplicable & Bad goal & Unsolved & Mean len & Max len & Time (s) & \# states \\
\midrule
\strips 50 states & .94 (.98) & .04 (.00) & .01 (.02) & .00 (.00) & 36 (36) & 81 (82) & .0126 (.0125) & 310 (295) \\
\strips 100 states & .69 (\textbf{1.0}) & .31 (.00) & .00 (.00) & .00 (.00) & 32 (35) & 71 (84) & .0122 (.0146) & 281 (331) \\
\strips atoms & .63 (.92) & .29 (.00) & .00 (.00) & .07 (.08) & 33 (46) & 77 (110) & .0144 (.0204) & 482 (758) \\
\midrule
SB 50 states & \textbf{1.0} (\textbf{1.0}) & .00 (.00) & .00 (.00) & .00 (.00) & 40 (40) & 90 (90) & .0082 (.0081) & 274 (274) \\
SB 100 states & \textbf{1.0} (\textbf{1.0}) & .00 (.00) & .00 (.00) & .00 (.00) & 36 (36) & 78 (78) & .0072 (.0072) & 230 (230) \\
SB atoms & \textbf{1.0} (\textbf{1.0}) & .00 (.00) & .00 (.00) & .00 (.00) & 40 (40) & 94 (94) & .0078 (.0077) & 247 (247) \\
\bottomrule
\end{tabular}

\vspace{0.1cm}

\begin{tabular}{@{}lrrrrrrrr@{}}
\toprule
 & \multicolumn{8}{c}{Maze 5x5} \\
\cmidrule(lr){2-9}
Model & Correct & Inapplicable & Bad goal & Unsolved & Mean len & Max len & Time (s) & \# states \\
\midrule
\strips 50 states & .63 (.70) & .15 (.12) & .00 (.00) & .21 (.18) & 7 (7) & 10 (10) & .0013 (.0013) & 9 (9) \\
\strips 100 states & .34 (.37) & .12 (.15) & .00 (.00) & .54 (.48) & 6 (6) & 10 (10) & .0013 (.0013) & 8 (8) \\
\strips atoms & .11 (.16) & .52 (.50) & .00 (.00) & .38 (.34) & 7 (7) & 9 (8) & .0022 (.0022) & 13 (13) \\
\midrule
SB 50 states & .90 (.90) & .00 (.00) & .00 (.00) & .10 (.10) & 8 (8) & 10 (10) & .0010 (.0009) & 9 (9) \\
SB 100 states & \textbf{1.0} (\textbf{1.0}) & .00 (.00) & .00 (.00) & .00 (.00) & 8 (8) & 11 (11) & .0010 (.0010) & 9 (9) \\
SB atoms & .98 (.98) & .00 (.00) & .00 (.00) & .02 (.02) & 8 (8) & 11 (11) & .0010 (.0010) & 9 (9) \\
\bottomrule
\end{tabular}

\vspace{0.1cm}

\begin{tabular}{@{}lrrrrrrrr@{}}
\toprule
 & \multicolumn{8}{c}{Maze 7x7} \\
\cmidrule(lr){2-9}
Model & Correct & Inapplicable & Bad goal & Unsolved & Mean len & Max len & Time (s) & \# states \\
\midrule
\strips 50 states & .44 (.50) & .51 (.44) & .00 (.00) & .05 (.06) & 13 (14) & 18 (19) & .0038 (.0036) & 18 (19) \\
\strips 100 states & .15 (.19) & .67 (.64) & .00 (.00) & .17 (.17) & 10 (10) & 17 (18) & .0038 (.0036) & 15 (15) \\
\strips atoms & .00 (.00) & .40 (.26) & .00 (.00) & .60 (.74) & --- & --- & --- & --- \\
\midrule
SB 50 states & .64 (.64) & .00 (.00) & .00 (.00) & .36 (.36) & 14 (14) & 19 (19) & .0024 (.0025) & 18 (18) \\
SB 100 states & .99 (.99) & .00 (.00) & .00 (.00) & .01 (.01) & 15 (15) & 20 (20) & .0028 (.0023) & 19 (19) \\
SB atoms & .98 (.98) & .00 (.00) & .00 (.00) & .02 (.02) & 16 (16) & 20 (20) & .0022 (.0023) & 20 (20) \\
\bottomrule
\end{tabular}

\vspace{0.1cm}

\begin{tabular}{@{}lrrrrrrrr@{}}
\toprule
 & \multicolumn{8}{c}{Logistics 3 cities} \\
\cmidrule(lr){2-9}
Model & Correct & Inapplicable & Bad goal & Unsolved & Mean len & Max len & Time (s) & \# states \\
\midrule
\strips 50 states & .95 (.97) & .01 (.03) & .00 (.00) & .03 (.00) & 11 (12) & 23 (23) & .0022 (.0023) & 127 (135) \\
\strips 100 states & .91 (.97) & .04 (.03) & .00 (.00) & .05 (.00) & 12 (12) & 28 (32) & .0023 (.0020) & 157 (144) \\
\strips atoms & .71 (.90) & .00 (.00) & .00 (.00) & .29 (.10) & 9 (9) & 18 (17) & .0015 (.0016) & 58 (62) \\
\midrule
SB 50 states & \textbf{1.0} (\textbf{1.0}) & .00 (.00) & .00 (.00) & .00 (.00) & 10 (10) & 20 (20) & .0016 (.0016) & 66 (66) \\
SB 100 states & \textbf{1.0} (\textbf{1.0}) & .00 (.00) & .00 (.00) & .00 (.00) & 10 (10) & 21 (21) & .0016 (.0016) & 65 (65) \\
SB atoms & \textbf{1.0} (\textbf{1.0}) & .00 (.00) & .00 (.00) & .00 (.00) & 10 (10) & 18 (18) & .0015 (.0015) & 62 (62) \\
\bottomrule
\end{tabular}

\vspace{0.1cm}

\begin{tabular}{@{}lrrrrrrrr@{}}
\toprule
 & \multicolumn{8}{c}{Logistics 5 cities} \\
\cmidrule(lr){2-9}
Model & Correct & Inapplicable & Bad goal & Unsolved & Mean len & Max len & Time (s) & \# states \\
\midrule
\strips 50 states & .54 (.73) & .26 (.12) & .00 (.00) & .20 (.15) & 27 (28) & 56 (65) & .1602 (.2464) & 8758 (13879) \\
\strips 100 states & .38 (.37) & .34 (.28) & .00 (.00) & .28 (.35) & 25 (25) & 43 (42) & .0917 (.0672) & 5197 (3824) \\
\strips atoms & .88 (\textbf{1.0}) & .00 (.00) & .00 (.00) & .12 (.00) & 23 (23) & 34 (34) & .0072 (.0070) & 325 (322) \\
\midrule
SB 50 states & \textbf{1.0} (\textbf{1.0}) & .00 (.00) & .00 (.00) & .00 (.00) & 23 (23) & 39 (39) & .0082 (.0073) & 330 (330) \\
SB 100 states & \textbf{1.0} (\textbf{1.0}) & .00 (.00) & .00 (.00) & .00 (.00) & 22 (22) & 36 (36) & .0067 (.0066) & 317 (317) \\
SB atoms & \textbf{1.0} (\textbf{1.0}) & .00 (.00) & .00 (.00) & .00 (.00) & 23 (23) & 34 (34) & .0071 (.0066) & 322 (322) \\
\bottomrule
\end{tabular}

\caption{\textbf{Extended planning results} (continued).}
\label{tab:extended_planning_part2}
\end{table*}